\documentclass[10pt,twocolumn,letterpaper]{article}

\usepackage{cvpr}
\usepackage{times}
\usepackage{epsfig}
\usepackage{graphicx}
\usepackage{amsmath}
\usepackage{amssymb}
\usepackage{algorithmicx}
\usepackage[ruled]{algorithm}
\usepackage{algpseudocode}
\usepackage{cite}
\usepackage{caption}
\usepackage{array}
\usepackage{marvosym}
\usepackage[noblocks]{authblk}

\usepackage[pagebackref=true,breaklinks=true,letterpaper=true,colorlinks,bookmarks=false]{hyperref}

\cvprfinalcopy % *** Uncomment this line for the final submission

\def\cvprPaperID{4380} % *** Enter the CVPR Paper ID here

\makeatletter
\renewcommand\normalsize{%
	\@setfontsize\normalsize\@xpt\@xiipt
	\abovedisplayskip 3\p@ \@plus2\p@ \@minus3\p@
	\abovedisplayshortskip \z@ \@plus2\p@
	\belowdisplayshortskip 3\p@ \@plus3\p@ \@minus5\p@
	\belowdisplayskip \abovedisplayskip
	\let\@listi\@listI}
\makeatother

\ifcvprfinal\fi
\begin{document}

%%%%%%%%% TITLE
\title{Neural Blind Deconvolution Using Deep Priors}

\author[1]{Dongwei Ren}
\author[2]{Kai Zhang}
\author[1]{Qilong Wang}
\author[1(\Letter)]{Qinghua Hu}
\author[2]{Wangmeng Zuo}

\affil[1]{Tianjin Key Lab of Machine Learning, College of Intelligence and Computing, Tianjin University, Tianjin, China}
\affil[2]{School of Computer Science and Technology, Harbin Institute of Technology, Harbin, China }
%\affil[3]{Tianjin Key Lab of Machine Learning, Tianjin, China}
\affil[ ]{\{csdren, huqinghua\}@tju.edu.cn}

\maketitle
\thispagestyle{empty}

%%%%%%%%% ABSTRACT
\begin{abstract}
	Blind deconvolution is a classical yet challenging low-level vision problem with many real-world applications.
	Traditional maximum a posterior (MAP) based methods rely heavily on fixed and handcrafted priors that certainly are insufficient in characterizing clean images and blur kernels, and usually adopt specially designed alternating minimization to avoid trivial solution.
	In contrast, existing deep motion deblurring networks learn from massive training images the mapping to clean image or blur kernel, but are limited in handling various complex and large size blur kernels.
	To connect MAP and deep models, we in this paper present two generative networks for respectively modeling the deep priors of clean image and blur kernel, and propose an unconstrained neural optimization solution to blind deconvolution.
	In particular, we adopt an asymmetric Autoencoder with skip connections for generating latent clean image, and a fully-connected network (FCN) for generating blur kernel.
	Moreover, the \emph{SoftMax} nonlinearity is applied to the output layer of FCN to meet the non-negative and equality constraints.
	The process of neural optimization can be explained as a kind of ``zero-shot" self-supervised learning of the generative networks, and thus our proposed method is dubbed SelfDeblur.
	Experimental results show that our SelfDeblur can achieve notable quantitative gains as well as more visually plausible deblurring results in comparison to state-of-the-art blind deconvolution methods on benchmark datasets and real-world blurry images.
	The source code is publicly available at \url{https://github.com/csdwren/SelfDeblur}. 
\end{abstract}

%%%%%%%%% BODY TEXT
\section{Introduction}

\begin{figure}[!tb]
	\footnotesize
	\centering
	\setlength{\tabcolsep}{1pt}
		\setlength{\abovecaptionskip}{0.cm}
	\setlength{\belowcaptionskip}{-0.cm}
	\begin{tabular}{cccccccccccccccccc}
		\includegraphics[width=.22\textwidth]{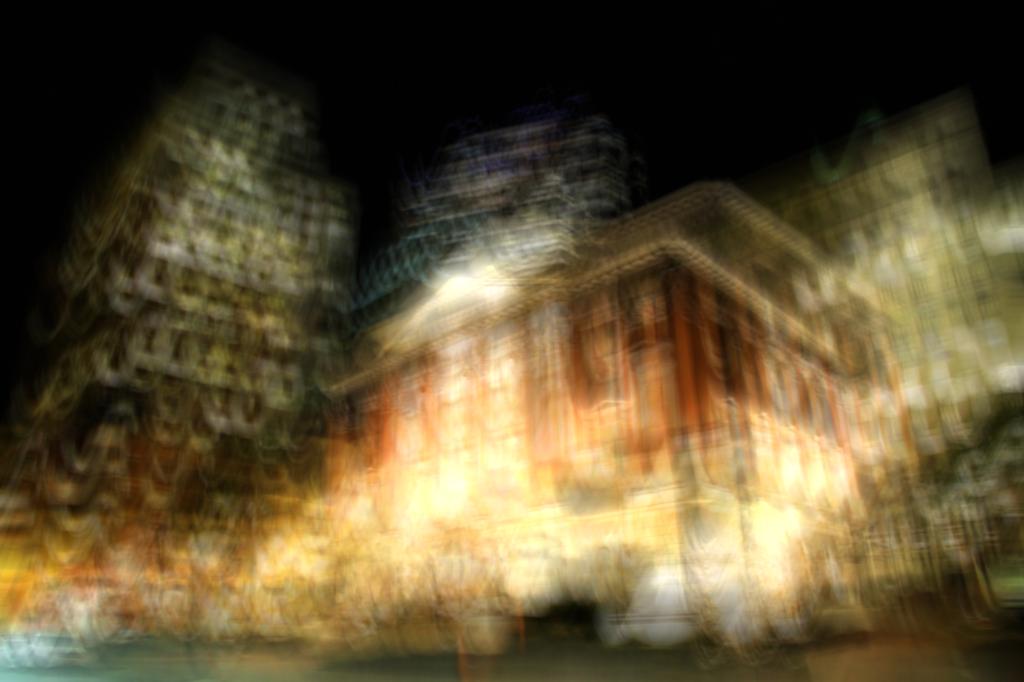} &
		\includegraphics[width=.22\textwidth]{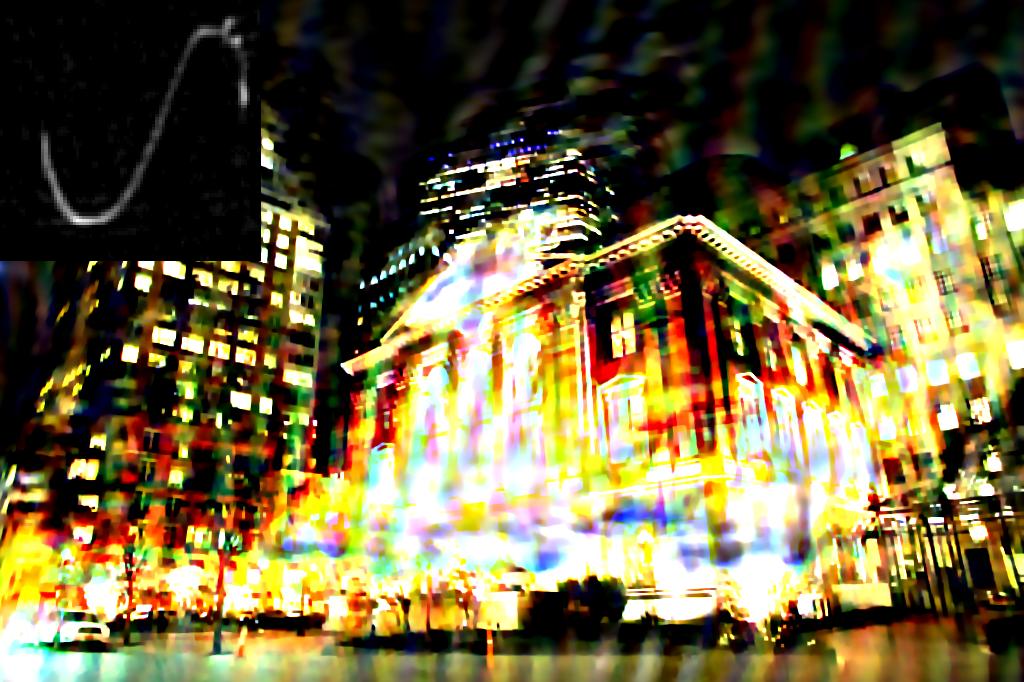} \\
		Blurry image & Xu \& Jia \cite{xu2010two}\\
		\includegraphics[width=.22\textwidth]{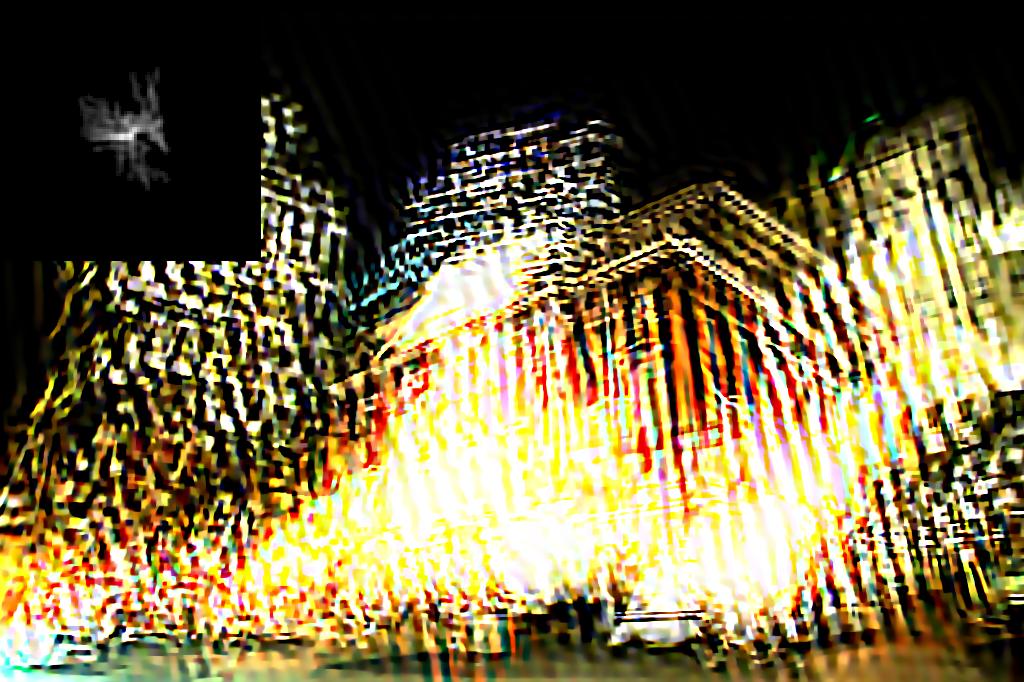} &
		\includegraphics[width=.22\textwidth]{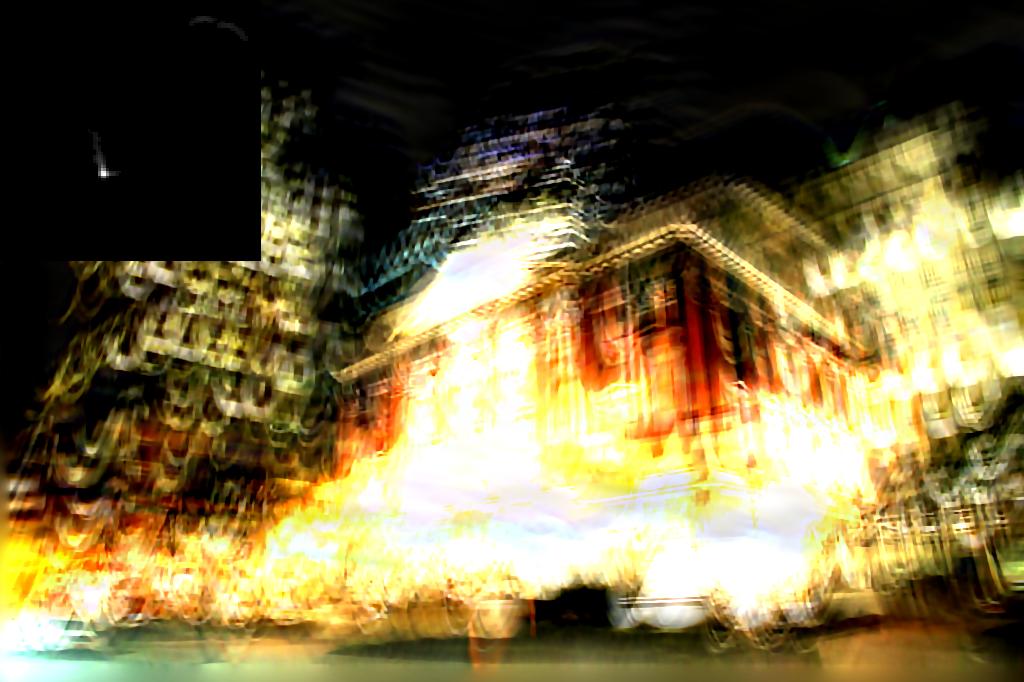} \\
		Pan-L0\cite{pan2017l_0} & Sun \etal \cite{sun2013edge}\\
		\includegraphics[width=.22\textwidth]{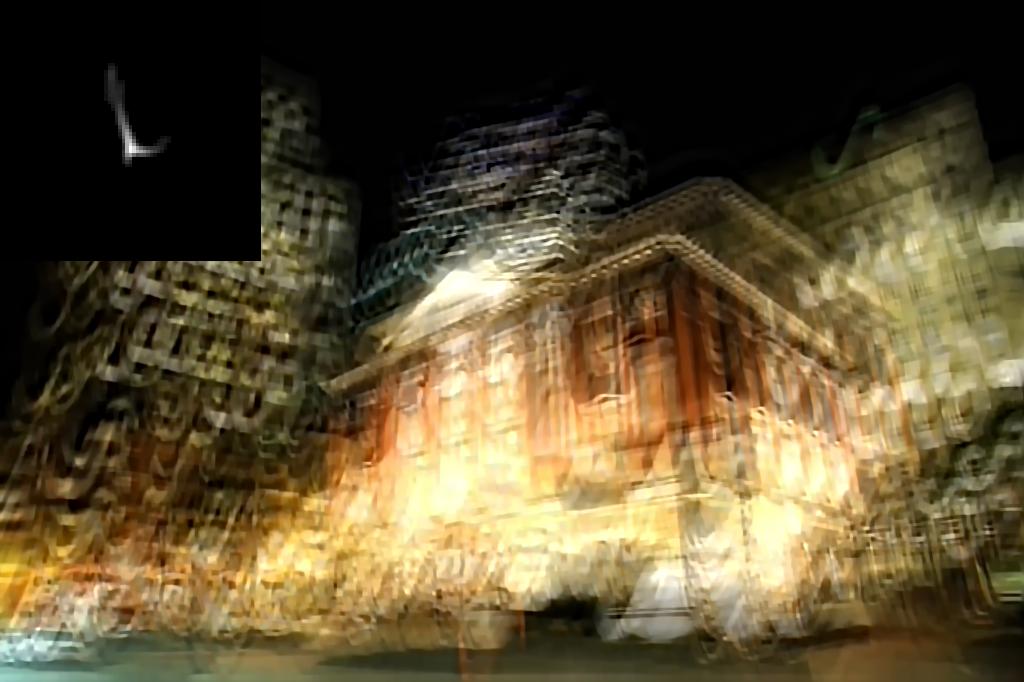} &
		\includegraphics[width=.22\textwidth]{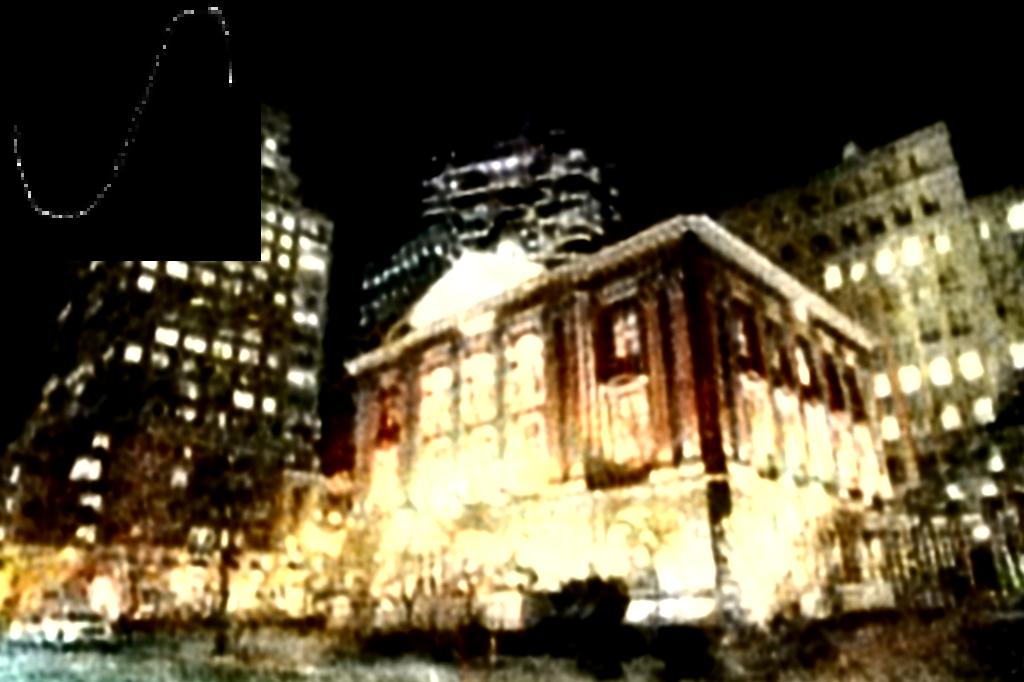} \\
		Pan-DCP\cite{pan2018deblurring} & SelfDeblur\\
	\end{tabular}
	\caption{Visual quality comparison on a severe blurry image.
%		SelfDeblur succeeds in estimating large size ($75\times75$) blur kernel, and generates visually satisfying latent clean image. 
	 }
	\label{fig:first example}
	\vspace{-0.15in}
\end{figure}

Camera shake during exposure inevitably yields blurry images and is a long-standing annoying issue in digital photography.
The removal of distortion from a blurry image, \ie, image deblurring, is a classical ill-posed problem in low-level vision and has received considerable research attention \cite{chan1998total,cho2009fast,levin2009understanding,pan2018learning,zuo2016learning,perrone2014total,pan2018deblurring,chen2019blind,jin2018normalized}.
When the blur kernel is spatially invariant, it is also known as blind deconvolution, where the blurry image $\mathbf{y}$ can be formulated as,
\begin{equation}\label{eq:convolution model}
\mathbf{y} = \mathbf{k} \otimes \mathbf{x} + \mathbf{n},
\end{equation}
where $\otimes$ denotes the 2D convolution operator, $\mathbf{x}$ is the latent clean image, $\mathbf{k}$ is the blur kernel, and $\mathbf{n}$ is the additive white Gaussian noise (AWGN) with noise level $\sigma$.
It can be seen that blind deconvolution should estimate both $\mathbf{k}$ and $\mathbf{x}$ from a blurry image $\bf y$, making it remain a very challenging problem after decades of studies.
%
%Blind deconvolution is a very challenging task, because both blur kernel $\mathbf{k}$ and latent clean image $\mathbf{x}$ should be estimated from the blurry image $\bf y$.

Most traditional blind deconvolution methods are based on the Maximum a Posterior (MAP) framework,
\begin{equation} \label{eq:map}
\begin{aligned}\footnotesize
\left(\mathbf{k},\mathbf{x}\right) & =
\arg \underset{\mathbf{x},\mathbf{k}}{\mathop{\max }}\,
\Pr \left(\mathbf{k}, \mathbf{x}|{\mathbf{y}} \right), \\
&= \arg \underset{\mathbf{x},\mathbf{k}}{\mathop{\max }} \Pr \left( {\mathbf{y}}|\mathbf{k},\mathbf{x} \right)
\Pr \left( \mathbf{x} \right)
\Pr \left( \mathbf{k} \right),
\end{aligned}
\end{equation}
where $\Pr \left( {\mathbf{y}}|\mathbf{k},\mathbf{x} \right)$ is the likelihood corresponding to the fidelity term, and $\Pr \left( \mathbf{x} \right)$ and $\Pr \left( \mathbf{k} \right)$ model the priors of clean image and blur kernel, respectively.
%
%In the past decades, the study of proper priors $\Pr \left( \mathbf{x} \right)$ and $\Pr \left( \mathbf{k} \right)$ has received extensive research attention.
%
%
%However, the success of MAP based methods relies heavily on the priors.
%
Although many priors have been suggested for $\mathbf{x}$ \cite{chan1998total,pan2017l_0,zuo2016learning,krishnan2011blind} and $\mathbf{k}$ \cite{zuo2016learning,pan2018deblurring,levin2009understanding,liu2014blind,sun2013edge,michaeli2014blind,ren2016image,pan2018deblurring,yan2017image}, they generally are handcrafted and certainly are insufficient in characterizing clean images and blur kernels.
Furthermore, the non-convexity of MAP based models also increases the difficulty of optimization.
%
%the constrains on $\mathbf{x}$ and $\mathbf{k}$ as in Eqn. \eqref{eq:blind deconvolution} are crucial for blur kernel estimation.
% \ie, pixels of image $\mathbf{x}$ should lie in the range $[0,1]$ and $\mathbf{k}$ has to meet the nonnegative normalization constraint.
%
%The constrained blind deconvolution is not trivial to solve, and
%
Levin \etal \cite{levin2009understanding} reveal that MAP-based methods may converge to trivial solution of delta kernel.
%
%\ie, ($\delta,\mathbf{y}$).
%
Perrone and Favaro \cite{perrone2014total} show that the success of existing methods can be attributed to some optimization details, \eg, projected alternating minimization and delayed normalization of $\mathbf{k}$.
%
%Once the blur kernel is estimated, non-blind deconvolution method is required to generate deblurring results with fine details.

\begin{figure*}[!htb]\small
	\centering
		\setlength{\abovecaptionskip}{0.cm}
	\setlength{\belowcaptionskip}{-0.cm}
	\includegraphics[width=0.8\textwidth]{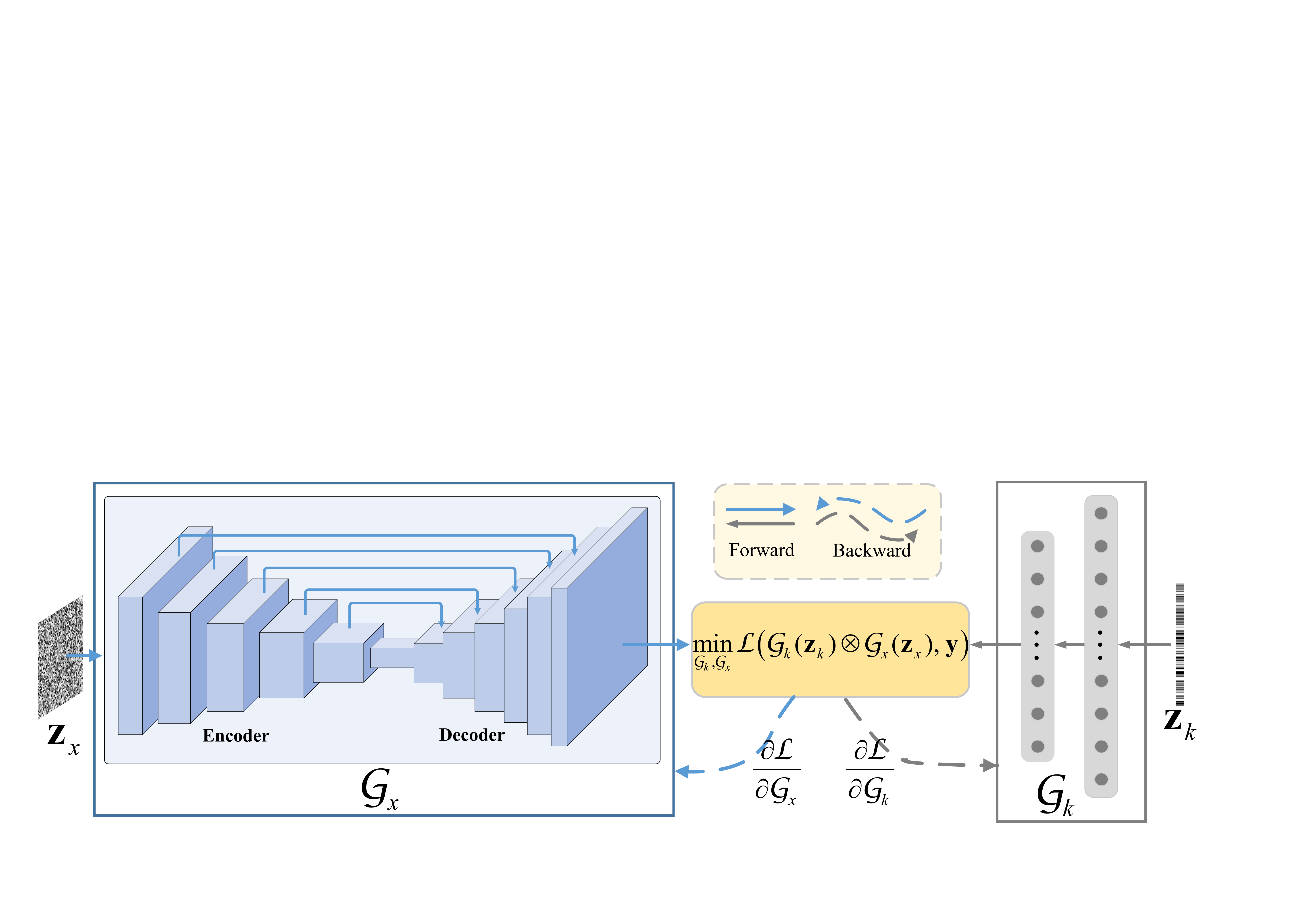}
	
	\caption{Illustration of our SelfDeblur method.
		The generative networks $\mathcal{G}_x$ and $\mathcal{G}_k$ are deployed to respectively capture the deep priors of blur kernel and latent clean image, and are trained using only the input blurry image.
		In particular, $\mathcal{G}_x$ is an asymmetric Autoencoder with skip connections, and the \emph{Sigmoid} nonlinearity is adopted to the output layer for constraining the pixels in $\mathbf{x}$ in the range $[0,1]$.
		$\mathcal{G}_k$ adopts a fully-connected network, where the \emph{SoftMax} nonlinearity is applied to the output layer for meeting the non-negative and equality constraints, and the 1D output of is reshaped to 2D blur kernel.
		%
		%The output of $\mathcal{G}_k$ is 1D signal, which is reshaped to 2D blur kernel.
		%
		%                                                                                Neural blind deconvolution is reformulated into unconstrained optimization problem by that (i) a \emph{Sigmoid} layer is adopted to constrain network output of $\mathcal{G}_x$ lying in the range $[0,1]$ and (ii) a \emph{SoftMax} layer is applied to the output of $\mathcal{G}_k$ to meet the non-negative and equality constraints of blur kernel, which can then be easily solved using either alternating minimization or joint minimization.
	}
	\label{fig:illustration}
	\vspace{-0.2in}
\end{figure*}

%%
%In the seminal work of \cite{chan1998total},  Chan \etal introduce total variation (TV) to model latent image prior, which motivates several variants based on gradient-based priors, \eg, $\ell_0$-norm \cite{pan2017l_0} and $\ell_p$-norm \cite{zuo2016learning}.
%%
%Levin \etal \cite{levin2009understanding} theoretically analyze that TV-based blind deconvolution and its variants would converge to the trivial solution \ie, ($\delta,\mathbf{y}$).
%%
%The success of existing MAP-based blind deconvolution methods can be attributed to some optimization details, \eg, projected alternating minimization algorithm, delayed normalization of blur kernel \cite{perrone2014total} and multi-scale implementation.
%%
%To avoid achieving the trivial solution, recent proposed blind deconvolution methods design specific regularization priors on $\bf x$ and $\bf k$.
%%
%As for $\Pr(\mathbf{x})$, gradient-based unnatural $\ell_0$-norm \cite{xu2013unnatural} and $\ell_1/\ell_2$-norm \cite{krishnan2011blind}, patch-based prior \cite{sun2013edge,michaeli2014blind}, low-rank prior \cite{ren2016image} and dark channel prior \cite{pan2018deblurring,yan2017image} are proposed.
%%
%For $\Pr(\mathbf{k})$, gradient sparsity priors \cite{zuo2016learning,pan2018deblurring,levin2009understanding} and the spectral prior \cite{liu2014blind} are specifically designed to model the distribution of blur kernel.
%%
%Once the blur kernel is estimated, non-blind deconvolution method is required to generate deblurring results with fine details.
%

Motivated by the unprecedented success of deep learning in low-level vision \cite{kim2016accurate,zhang2017beyond,kim2016deeply,tai2017image,liu2019convergence}, some attempts have also been made to solve blind deconvolution using deep convolutional neural networks (CNNs).
Given the training set, deep CNNs can either be used to extract features to facilitate blur kernel estimation \cite{schuler2016learning,chakrabarti2016neural}, or be deployed to learn the direct mapping to clean image for motion deblurring \cite{zhang2018dynamic,nah2017deep,tao2018scale,gao2019dynamic}.
However, these methods do not succeed in handling various complex and large size blur kernels in blind deconvolution.
%
%Without estimating blur kernels, these deep deblurring networks directly map a blurry image to its latent clean image.
%
%These deep motion deblurring methods, however, do not succeed in addressing blind deconvolution, due to the more severe ill-posedness caused by diverse blur kernels.
%
Recently, Ulyanov \etal~\cite{ulyanov2018deep} suggest the deep image prior (DIP) framework, which adopts the structure of a DIP generator network to capture low-level image statistics and shows powerful ability in image denoising, super-resolution, inpainting, \etc.
Subsequently, Gandelsman \etal~\cite{gandelsman2019double} combine multiple DIPs (\ie, Double-DIP) for multi-task layer decomposition such as image dehazing and transparency separation.
However, Double-DIP cannot be directly applied to solve blind deconvolution due to that the DIP network is designed to generate natural images and is limited to capture the prior of blur kernels.

%Motivated by the deep image prior (DIP) \cite{ulyanov2018deep} for image restoration,
In this paper, we propose a novel neural optimization solution to blind deconvolution.
Motivated by the DIP network~\cite{ulyanov2018deep}, an image generator network $\mathcal{G}_x$, \ie, an asymmetric Autoencoder with skip connections, is deployed to capture the statistics of latent clean image.
Nonetheless, image generator network cannot well characterize the prior on blur kernel.
Instead, we adopt a fully-connected network (FCN) $\mathcal{G}_k$ to model the prior of blur kernel.
Furthermore, the \emph{SoftMax} nonlinearity is deployed to the output layer of $\mathcal{G}_k$, and the non-negative and equality constraints on blur kernel can then be naturally satisfied.
By fixing the network structures ($\mathcal{G}_k$ and $\mathcal{G}_x$) and inputs ($\mathbf{z}_k$ and $\mathbf{z}_x$) sampled from uniform distribution, blind deconvolution is thus formulated as an unconstrained neural optimization on network parameters of $\mathcal{G}_k$ and $\mathcal{G}_x$.
As illustrated in Fig. \ref{fig:illustration}, given a blurry image $\bf y$, the optimization process can also be explained as a kind of ``zero-shot'' self-supervised learning \cite{shocher2018zero} of $\mathcal{G}_k$ and $\mathcal{G}_x$, and our proposed method is dubbed SelfDeblur.
%
%where deep priors of blur kernel and latent clean image can be specifically tailored to the given input blurry image itself (SelfDeblur).
%
%As shown in Fig. \ref{fig:illustration}, we employ two generative networks to model the priors of latent clean image ($\mathcal{G}_x$) and blur kernel ($\mathcal{G}_k$), respectively.
%
%In particular, $\mathbf{z}_k$ and $\mathbf{z}_x$ are with the uniform distribution, and the regularization priors of blur kernel and latent clean image can be modeled by the sufficient generative models $\mathcal{G}_k$ and $\mathcal{G}_x$.
%
%The proposed neural blind deconvolution is further reformulated into unconstrained optimization problem by absorbing the constraints into deep generative networks, and
%
Even though SelfDeblur can be optimized with either alternating optimization or joint optimization, our empirical study shows that the latter performs better in most cases.

Experiments are conducted on two widely used benchmarks \cite{levin2009understanding,lai2016comparative} as well as real-world blurry images to evaluate our SelfDeblur.
Fig. \ref{fig:first example} shows the deblurring results on a severe real-world blurry image.
While the competing methods either fail to estimate large size blur kernels or suffer from ringing effects, our SelfDeblur succeed in estimating the blur kernel and generating visually favorable deblurring image.
In comparison to the state-of-the-art methods, our SelfDeblur can achieve notable quantitative performance gains and performs favorably in generating visually plausible deblurring results.
It is worth noting that our SelfDeblur can both estimate blur kernel and generate latent clean image with satisfying visual quality, making the subsequent non-blind deconvolution not a compulsory choice.

Our contributions are summarized as follows:
\begin{itemize}
	\vspace{-0.06in}
	\item
	A neural blind deconvolution method, \ie, SelfDeblur, is proposed, where DIP and FCN are respectively introduced to capture the priors of clean image and blur kernel. And the \emph{SoftMax} nonlinearity is applied to the output layer of FCN to meet the non-negative and equality constraints.
	\vspace{-0.06in}
	\item
	The joint optimization algorithm is suggested to solve the unconstrained neural blind deconvolution model for both estimating blur kernel and generating latent clean image, making the non-blind deconvolution not a compulsory choice for our SelfDeblur.
	%	
	%	\item
	%	The deep priors not only contribute to blur kernel estimation but also succeed in generating latent clean image with fine details, making the subsequent non-blind deconvolution be optional.
	%	
	\vspace{-0.06in}
	\item
	Extensive experiments show that our SelfDeblur performs favorably against the existing MAP-based methods in terms of quantitative and qualitative evaluation.
	To our best knowledge, SelfDeblur makes the first attempt of applying deep learning to yield state-of-the-art blind deconvolution performance.
\end{itemize}

\section{Related Work}
In this section, we briefly survey the relevant works including optimization-based blind deconvolution and deep learning based blind deblurring methods.

\subsection{Optimization-based Blind Deconvolution}

Traditional optimization-based blind deconvolution methods can be further categorized into two groups, \ie, Variational Bayes (VB)-based and MAP-based methods.
VB-based method \cite{levin2011efficient} is theoretically promising, but is with heavy computational cost. %and the performance is usually inferior to MAP-based methods.
As for the MAP-based methods, many priors have been suggested for modeling clean images and blur kernels.
In the seminal work of \cite{chan1998total}, Chan \etal introduce the total variation (TV) regularization to model latent clean image in blind deconvolution, and motivates several variants based on gradient-based priors, \eg, $\ell_0$-norm \cite{pan2017l_0} and $\ell_p$-norm \cite{zuo2016learning}.
Other specifically designed regularizations, \eg, $\ell_1/\ell_2$-norm \cite{krishnan2011blind}, patch-based prior \cite{sun2013edge,michaeli2014blind}, low-rank prior \cite{ren2016image} and dark channel prior \cite{pan2018deblurring,yan2017image} have also been proposed to identify and preserve salient edges for benefiting blur kernel estimation.
Recently, a discriminative prior \cite{li2018learning} is presented to distinguish the clean image from a blurry one, but still heavily relies on $\ell_0$-norm regularizer for attaining state-of-the-art performance.
As for $\Pr(\mathbf{k})$, gradient sparsity priors \cite{zuo2016learning,pan2018deblurring,levin2009understanding} and spectral prior \cite{liu2014blind} are usually adopted.
In order to solve the MAP-based model, several tricks have been introduced to the projected alternating minimization algorithm, including delayed normalization of blur kernel~\cite{perrone2014total}, multi-scale implementation \cite{krishnan2011blind} and time-varying parameters \cite{zuo2016learning}.

%from gradient sparsity-based regularization priors (\eg, TV \cite{chan1998total}, $\ell_0$-norm and $\ell_p$-norm) to specific regularization forms (\eg, ${\ell_1}/{\ell_2}$-norm \cite{krishnan2011blind}, patch prior \cite{sun2013edge}, low-rank prior \cite{ren2016image} and dark channel prior \cite{pan2018deblurring,yan2017image}),
%the success of blur kernel estimation is based on the widely known fact that salient edges preservation \cite{cho2009fast,xu2010two} in latent image $\mathbf{x}$ is the crucial step in alternating optimization algorithms.
%
%To this end, the constrained blind deconvolution should be solved using projected alternating minimization algorithm with some optimization details (\eg, delayed normalization of blur kernel, multi-scale implementation and time-varying parameters \cite{zuo2016learning}) to gradually select proper salient edges.
%
%Most recently, a discriminative prior \cite{li2018learning} to distinguish the clean image from a blurry image is learned for blur kernel estimation, but still heavily rely on $\ell_0$-norm for state-of-the-art performance.
%

After blur kernel estimation, non-blind deconvolution is required to recover the latent clean image with fine texture details \cite{krishnan2011blind,pan2017l_0,pan2018deblurring,sun2013edge,xu2013unnatural,liu2014blind}.
%
%the latent image $\mathbf{x}$ by these state-of-the-art blind deconvolution methods \cite{krishnan2011blind,pan2017l_0,pan2018deblurring,sun2013edge,xu2013unnatural,liu2014blind} is usually over-smoothed without texture details due to the salient edge preservation steps, and thus another non-blind deconvolution method should be finally adopted to generate clean image with finer texture details.
%
Thus, the priors should favor natural images, \eg, hyper-Laplacian \cite{krishnan2009fast}, GMM \cite{zoran2011learning}, non-local similarity \cite{dong2013nonlocally}, \eg, RTF \cite{schmidt2013discriminative}, CSF \cite{schmidt2014shrinkage} and CNN \cite{kruse2017learning,zhang2017learning}, which are quite different from those used in blur kernel estimation.
Our SelfDeblur can be regarded as a special MAP-based method, but two generative networks, \ie, DIP and FCN, are adopted to respectively capture the deep priors of clean image and blur kernel.
Moreover, the joint optimization algorithm is effective to estimate blur kernel and generate clean image, making non-blind deconvolution not a compulsory choice for SelfDeblur.

%the neural blind deconvolution can be easily addressed using joint optimization algorithm, under which the deep priors provide a unified solution to simultaneously estimate blur kernel and generate clean image with fine details.
%The non-blind deconvolution is optional but not necessary for our SelfDeblur.
%

%
%Other specifically designed regularizations, \eg, $\ell_1/\ell_2$-norm \cite{krishnan2011blind}, patch-based prior \cite{sun2013edge,michaeli2014blind}, low-rank prior \cite{ren2016image} and dark channel prior \cite{pan2018deblurring,yan2017image} have also been proposed for better blur kernel estimation.

\subsection{Deep Learning in Image Deblurring}

Many studies have been given to apply deep learning (DL) to blind deblurring.
For example, DL can be used to help the learning of mapping to blur kernel.
By imitating the alternating minimization steps in optimization-based methods, Schuler \etal \cite{schuler2016learning} design the deep network architectures for blur kernel estimation.
%
%This work suggests that the learned deep features are different from salient edges.
%
By studying the spectral property of blurry images, deep CNN is suggested to predict the Fourier coefficients \cite{chakrabarti2016neural}, which can then be projected to estimate blur kernel.
In \cite{sun2015learning}, CNN is used to predict the parametric blur kernels for motion blurry images.
%
%However, these deep blind deconvolution networks are limited in handling large size blur kernels.

For dynamic scene deblurring, deep CNNs have been developed to learn the direct mapping to latent clean image \cite{zhang2018dynamic,nah2017deep,tao2018scale,sun2015learning,kupyn2018deblurgan}.
Motivated by the multi-scale strategy in blind deconvolution, multi-scale CNN \cite{nah2017deep} and scale-recurrent network \cite{tao2018scale} are proposed to directly estimate the latent clean image from the blurry image.
The adversarial loss is also introduced for better recovery of texture details in motion deblurring \cite{kupyn2018deblurgan}.
Besides, by exploiting the temporal information between adjacent frames, deep networks have also been applied to video motion deblurring \cite{nah2019recurrent,hyun2017online,pan2017simultaneous}.
However, due to the severe ill-posedness caused by large size and complex blur kernels, existing DL-based methods still cannot outperform traditional optimization-based ones for blind deconvolution.

Recently, DIP~\cite{ulyanov2018deep} and Double-DIP~\cite{gandelsman2019double} have been introduced to capture image statistics, and have been deployed to many low-level vision tasks such as super-resolution, inpainting, dehazing, transparency separation, \etc.
Nonetheless, the DIP network is limited in capturing the prior of blur kernels, and Double-DIP still performs poorly for blind deconvolution.
To the best of our knowledge, our SelfDeblur makes the first attempt of applying deep networks to yield state-of-the-art blind deconvolution performance.

%Albeit superior quantitative metrics on synthetic datasets \cite{nah2017deep}, these motion deblurring networks cannot succeed in addressing blind deconvolution, because
%the blur kernels are so diverse and complex that cannot be modeled using one CNN.
%
%Different from previous DL-based methods, our SelfDeblur jointly optimizes two generative networks to effectively solve the unconstrained neural blind deconvolution given a blurry image.

%In \cite{schuler2016learning}, several CNNs are stacked to alternatively extract features and estimate blur kernel, by imitating the alternating optimization steps in conventional blind deconvolution methods.
%
%In \cite{chakrabarti2016neural}, CNN is used to predict the Fourier coefficients of blur kernel, from which the spectral property of blur kernel can be studied.
%
%However, these DL-based methods fail in handling large size blur kernels.
%
%On the other hand, deep CNN-based methods have achieved notable performance improvements in motion deblurring \cite{zhang2018dynamic,nah2017deep,tao2018scale}.
%
%Without estimating blur kernels, these deep deblurring networks directly map a blurry image to its latent clean image.
%
%These deep motion deblurring methods, however, do not succeed in addressing blind deconvolution, due to the more severe ill-posedness caused by diverse blur kernels.

\section{Proposed Method}
In this section, we first introduce the general formulation of MAP-based blind deconvolution, and then present our proposed neural blind deconvolution model as well as the joint optimization algorithm.

\subsection{MAP-based Blind Deconvolution Formulation}
According to Eqn.~\eqref{eq:convolution model}, we define the fidelity term as $ -\log\left( \Pr \left( {\mathbf{y}}|\mathbf{k},\mathbf{x} \right)\right) = \|\mathbf{k}\otimes\mathbf{x}-\mathbf{y}\|^2$.
And we further introduce two regularization terms $-\log(\Pr(\mathbf{x})) = \phi(\mathbf{x})$ and $-\log(\Pr(\mathbf{k})) = \varphi(\mathbf{k})$ for modeling the priors on latent clean image and blur kernel, respectively.
The MAP-based blind deconvolution model in Eqn. \eqref{eq:map} can then be reformulated as,
%
%According to Eqn. \eqref{eq:map}, blind deconvolution can be formulated as the following minimization problem,
%
\begin{equation}\label{eq:blind deconvolution}
\begin{aligned}
(&\mathbf{x},\mathbf{k})=\arg  \underset{(\mathbf{x},\mathbf{k})}{\min} \| \mathbf{k}\otimes \mathbf{x} - \mathbf{y} \|^2 + \lambda\phi(\mathbf{x}) + \tau\varphi(\mathbf{k})\\
& {s.t.} \ 0\leq x_i\leq 1, \forall i,   k_j \geq 0, \sum\nolimits_{j}{k_j} = 1, \forall j,\\
\end{aligned}
\end{equation}
where $\lambda$ and $\tau$ are trade-off regularization parameters.
%
%$\mathcal{D}$ is the fidelity term, $\phi(\mathbf{x}) = -\log(\Pr(\mathbf{x}))$ and $\varphi(\mathbf{k}) = -\log(\Pr(\mathbf{k}))$ are regularization terms for latent clean image and blur kernel, respectively.
%
Besides the two regularization terms, we further introduce the non-negative and equality constraints for blur kernel $\mathbf{k}$ \cite{sun2013edge,zuo2016learning,pan2018deblurring,perrone2014total}, and the pixels in $\bf x$ are also constrained to the range $[0,1]$.

Under the MAP-based framework, many fixed and handcrafted regularization terms have been presented for latent clean image and blur kernel \cite{zuo2016learning,pan2018deblurring,liu2014blind,sun2013edge,michaeli2014blind,ren2016image,pan2018deblurring,yan2017image}.
To solve the model in Eqn.~\eqref{eq:blind deconvolution}, projected alternating minimization is generally adopted, but several optimization details, \eg, delayed normalization~\cite{perrone2014total} and multi-scale implementation~\cite{krishnan2011blind}, are also crucial to the success of blind deconvolution.
Moreover, once the estimated blur kernel $\mathbf{\hat{k}}$ is obtained by solving Eqn.~\eqref{eq:blind deconvolution}, another non-blind deconvolution usually is required to generate final deblurring result,
\begin{equation}\label{eq:nonblind deconvolution}
\mathbf{x} \!=\! \arg  \underset{\mathbf{x}}{\min} \| \mathbf{\hat{k}} \otimes \mathbf{x} \!-\! \mathbf{y} \|^2 \!+\! \lambda \mathcal{R}(\mathbf{x}),
\end{equation}
where $\mathcal{R}(\mathbf{x})$ is a regularizer to capture natural image statistics and is quite different from $\phi(\mathbf{x})$.

%Since it is extremely hard to jointly optimize $\mathbf{k}$ and $\mathbf{x}$ in the constrained optimization problem Eqn. \eqref{eq:blind deconvolution}, existing blind deconvolution methods adopt projected alternating optimization, \ie, optimizing $\mathbf{x}$-subproblem by fixing $\mathbf{k}$, and vice versa.
%
%Then $\mathbf{k}$ and $\mathbf{x}$ are projected to the constrainted domains.
%
%Moreover, in practical implementions \cite{sun2013edge,pan2017l_0,pan2018deblurring}, the non-negative constraint of $\mathbf{k}$ is not strictly met during blur kernel estimation, but a hard threshold operation (\ie, $k_i \geq \epsilon >0$) is actually adopted to eliminate noises in blur kernel.
%
%And the time-varying parameters $\lambda$ and $\tau$ should be carefully tuned to properly identify and preserve salient edges in intermediate latent images.
%
%To sum up, the success of existing blind deconvolution methods comes from these optimization details \cite{perrone2014total}.
%
%Moreover, another non-blind deconvolution method should be performed to generate final latent clean images with visually plausible textures, after the blur kernel is estimated.
%

\subsection{Neural Blind Deconvolution}
Motivated by the success of DIP \cite{ulyanov2018deep} and Double-DIP \cite{gandelsman2019double}, we suggest the neural blind deconvolution model by adopting generative networks $\mathcal{G}_x$ and $\mathcal{G}_k$ to capture the priors of $\bf x$ and $\bf k$.
By substituting $\bf x$ and $\bf k$ with $\mathcal{G}_x$ and $\mathcal{G}_k$ and removing the regularization terms $\phi(\mathbf{x})$ and $\varphi(\mathbf{k})$, the neural blind deconvolution can be formulated as,
\begin{equation}\label{eq:self learn_v0}
\begin{aligned}
& \underset{(\mathcal{G}_x,\mathcal{G}_k)}{\min} \| \mathcal{G}_k(\mathbf{z}_k) \otimes \mathcal{G}_x(\mathbf{z}_x) - \mathbf{y} \|^2 \\
&s.t. \ \ 0\leq (\mathcal{G}_x(\mathbf{z}_x))_i \leq 1, \forall i, \\
& \ \ \ \ \ \ \ (\mathcal{G}_k(\mathbf{z}_k))_j \geq 0,  \sum\nolimits_{j}{(\mathcal{G}_k(\mathbf{z}_k))_j} = 1, \forall j,\\
\end{aligned}
\end{equation}
where $\mathbf{z}_x$ and $\mathbf{z}_k$ are sampled from the uniform distribution, $(\cdot)_i$ and $(\cdot)_j$ denote the $i$-th and $j$-th elements.
We note that $\mathbf{z}_k$ is 1D vector, and $\mathcal{G}_k(\mathbf{z}_k)$ is reshaped to obtain 2D matrix of blur kernel.
%All the other bold letters denote  

%Motivated by DIP \cite{ulyanov2018deep} and Double-DIP \cite{gandelsman2019double}, the deep networks have sufficient modeling capacity to serve as $\mathcal{G}_x$ and $\mathcal{G}_k$ to learn the personalized priors of $\bf x$ and $\bf k$.
%
However, there remain several issues to be addressed with neural blind deconvolution.
(i) The DIP network \cite{ulyanov2018deep} is designed to capture low-level image statistics and is limited in capturing the prior of blur kernels.
As a result, we empirically find that Double-DIP~\cite{gandelsman2019double} performs poorly for blind deconvolution (see the results in Sec.~\ref{sec:experiment architecture}).
(ii) Due to the non-negative and equality constraints, the resulting model in Eqn.~\eqref{eq:self learn_v0} is a constrained neural optimization problem and is difficult to optimize.
(iii) Although the generative networks $\mathcal{G}_x$ and $\mathcal{G}_k$ present high impedance to image noise, the denoising performance of DIP heavily relies on the additional averaging over last iterations and different optimization runs \cite{ulyanov2018deep}.
Such heuristic solutions, however, both bring more computational cost and cannot be directly borrowed to handle blurry and noisy images.

In the following, we present our solution to address the issues (i)\&(ii) by designing proper generative networks $\mathcal{G}_x$ and $\mathcal{G}_k$.
As for the issue (iii), we introduce an extra TV regularizer and a regularization parameter to explicitly consider noise level in the neural blind deconvolution model.

%the neural blind deconvolution into unconstrained optimization problem, and describe our developed generative networks for $\mathcal{G}_x$ and $\mathcal{G}_k$.

%\subsubsection{Unconstrained Neural Blind Deconvolution}
%By absorbing the constraints of $\bf x$ and $\bf k$ into generative networks $\mathcal{G}_x$ and $\mathcal{G}_k$ respectively, blind deconvolution can be reformulated as an unconstrained neural optimization problem.
%

\noindent \emph{\textbf{Generative Network $\mathcal{G}_x$.}}
The latent clean images usually contain salient structures and rich textures, which requires the generative network $\mathcal{G}_x$ to have sufficient modeling capacity.
Fortunately, since the introduction of generative adversarial network~\cite{goodfellow2014generative}, dramatic progress has been made in generating high quality natural images \cite{ulyanov2018deep}.
For modeling $\mathbf{x}$, we adopt a DIP network, \ie, the asymmetric Autoencoder \cite{ronneberger2015unet} with skip connections in \cite{ulyanov2018deep}, to serve as $\mathcal{G}_x$.
As shown in Fig.~\ref{fig:illustration}, the first 5 layers of encoder are skip connected to the last 5 layers of decoder.
Finally, a convolutional output layer is used to generate latent clean image.
To meet the range constraint for $\mathbf{x}$, the \emph{Sigmoid} nonlinearity is applied to the output layer.
%
%we note that the \emph{Sigmoid} nonlinearity is applied to the output layer, constraining the latent image lying in the range $[0,1]$.
%
Please refer to the supplementary file for more architecture details of $\mathcal{G}_x$.

%As for the range constraint of $\mathbf{x}$, it is straightfowrad to see that the \emph{Sigmoid} nonlinearity can be adopted to the output layer of $\mathcal{G}_x$ to project $\bf x$ lying in the pixel range $[0,1]$.
%

\noindent  \emph{\textbf{Generative Network $\mathcal{G}_k$.}}
On the one hand, the DIP network \cite{ulyanov2018deep} is designed to capture the statistics of natural image but performs limited in modeling the prior of blur kernel.
On the other hand, blur kernel $\mathbf{k}$ generally contains much fewer information than latent clean image $\mathbf{x}$, and can be well generated by simpler generative network.
Thus, we simply adopt a fully-connected network (FCN) to serve as $\mathcal{G}_k$.
As shown in Fig.~\ref{fig:illustration}, the FCN $\mathcal{G}_k$ takes a 1D noise $\mathbf{z}_k$ with 200 dimensions as input, and has a hidden layer of 1,000 nodes and an output layer of $K^2$ nodes.
To guarantee the non-negative and equalitly constraints can be always satisfied, the \emph{SoftMax} nonlinearity is applied to the output layer of $\mathcal{G}_k$.
Finally, the 1D output of $K^2$ entries is reshaped to a 2D $K \times K$ blur kernel.
Please refer to Suppl. for more architecture details of $\mathcal{G}_k$.

%The 2D blur kernel can be obtained by reshaping the 1D output of $\mathcal{G}_k$ to blur kernel size.
%
%As shown in Fig. \ref{fig:illustration}, a 1D noise $\mathbf{z}_k$ with 200 dimensions are fully connected to a hidden layer with 1,000 nodes, which are then connected to the output layer with nodes equal to the pixel numbers of blur kernel.
%
%Finally, the \emph{SoftMax} nonlinearity is applied to the output layer of $\mathcal{G}_k$ to guarantee that the non-negative and equalitly constraints of blur kernel in Eqn. \eqref{eq:self learn} is always satisfied.
%
%More architecture details of the network $\mathcal{G}_k$ can be found in the supplementary material.

%As for the non-negative and equality constraints of $\mathbf{k}$, it is interesting to see that the \emph{SoftMax} function can ensure that each value of $\bf k$ is non-negative and its sum equals 1.
%Thus, a \emph{SoftMax} layer is applied to the output of $\mathcal{G}_k$ to meet the non-negative and equality constraints of $\bf k$.
%

\noindent {\emph{\textbf{Unconstrained Neural Blind Deconvolution with TV Regularization.}}}
With the above generative networks $\mathcal{G}_x$ and $\mathcal{G}_k$, we can formulate neural blind deconvolution into an unconstrained optimization form.
However, the resulting model is irrelevant with the noise level, making it perform poorly on blurry images with non-negligible noise.
To address this issue, we combine both $\mathcal{G}_x$ and TV regularization to capture image priors, and our neural blind deconvolution model can then be written as,
\begin{equation}\label{eq:selfdeblur function}
\underset{\mathcal{G}_k,\mathcal{G}_x}{\min} \|\mathcal{G}_k(\mathbf{z}_k)\otimes \mathcal{G}_x(\mathbf{z}_x)\! -\!\mathbf{y}\|^2 \!+\! \lambda \text{TV}(\mathcal{G}_x(\mathbf{z}_x)),
\end{equation}
where $\lambda$ denotes the regularization parameter controlled by noise level $\sigma$.
Albeit the generative network $\mathcal{G}_x$ is more powerful, the incorporation of $\mathcal{G}_x$ and another image prior generally is beneficial to deconvolution performance.
Moreover, the introduction of the noise level related regularization parameter $\lambda$ can greatly improve the robustness in handling blurry images with various noise levels.
In particular, we emperically set $\lambda=0.1\times\sigma$ in our implementation, and the noise level $\sigma$ can be estimated using \cite{zoran2009scale}.

\vspace{-0.05in}
\subsection{Optimization Algorithm}
\vspace{-0.05in}
The optimization process of Eqn.~\eqref{eq:selfdeblur function} can be explained as a kind of "zero-shot" self-supervised learning \cite{shocher2018zero}, where the generative networks $\mathcal{G}_k$ and $\mathcal{G}_x$ are trained using only a test image (\ie, blurry image $\bf y$) and no ground-truth clean image is available.
Thus, our method is dubbed SelfDeblur.
%
%The proposed neural blind deconvolution Eqn. \eqref{eq:selfdeblur function} can be solved using gradient descent algorithm.
%
In the following, we present two algorithms for SelfDeblur, \ie, alternating optimization and joint optimization.

\vspace{0.05in}
\noindent\emph{\textbf{{Alternating Optimization.}}}
Analogous to the alternating minimization steps in traditional blind deconvolution \cite{zuo2016learning,chan1998total,pan2018deblurring,pan2017l_0,sun2013edge}, the network parameters of $\mathcal{G}_k$ and $\mathcal{G}_x$ can also be optimized in an alternating manner.
As summarized in Algorithm \ref{algm:self deblur ADM}, the parameters of $\mathcal{G}_k$ are updated via the ADAM \cite{kingma2014adam} by fixing $\mathcal{G}_x$, and vice versa.
In particular, the gradient w.r.t. either $\mathcal{G}_x$ or $\mathcal{G}_k$ can be derived using automatic differentiation \cite{paszke2017automatic}.

\vspace{0.05in}
\noindent\emph{\textbf{{Joint Optimization.}}}
In traditional MAP-based framework, alternating minimization allows the use of projection operator to handle non-negative and equality constraints and the modification of optimization details to avoid trivial solution, and thus has been widely adopted.
As for our neural blind deconvolution, the model in Eqn.~\eqref{eq:selfdeblur function} is unconstrained optimization, and the powerful modeling capacity of $\mathcal{G}_k$ and $\mathcal{G}_x$ is beneficial to avoid trivial delta kernel solution.
We also note that the unconstrained neural blind deconvolution is highly non-convex, and alternating optimization may get stuck at saddle points \cite{tseng2001convergence}.
Thus, joint optimization is more prefered than alternating optimization for SelfDeblur.
Using the automatic differentiation techniques \cite{paszke2017automatic}, the gradients w.r.t. $\mathcal{G}_k$ and $\mathcal{G}_x$ can be derived.
%
%It is also feasible to jointly compute the gradients of ojective function w.r.t. the parameters of $\mathcal{G}_k$ and $\mathcal{G}_x$, thanks to the automatic differentiation techniques \cite{paszke2017automatic}.
%
Algorithm \ref{algm:self deblur} summarizes the joint optimization algorithm, where the parameters of $\mathcal{G}_k$ and $\mathcal{G}_x$ can be jointly updated using the ADAM algorithm.
%
%We note that the unconstrained neural blind deconvolution Eqn. \eqref{eq:selfdeblur function} is highly non-convex, and alternating optimization algorithm is potential to pay a "price" \ie, getting stuck at a saddle point \cite{tseng2001convergence}.
%
Our empirical study in Sec.~\ref{sec:experiment joint} also shows that joint optimization usually converges to better solutions than alternating optimization.

%Thus, it is expected to converge to better solutions by the joint optimization of SelfDeblur, which is emperically supported by the experimental results .

Both alternating optimization and joint optimization algorithms are stopped when reaching $T$ iterations.
Then, the estimated blur kernel and latent clean image can be generated using $\mathbf{k}=\mathcal{G}^T_k(\mathbf{z}_k)$ and $\mathbf{x} = \mathcal{G}^T_x(\mathbf{z}_x)$, respectively.
Benefited from the modeling capacity of $\mathcal{G}_x(\mathbf{z}_x)$, the estimated  $\bf x$ is with visually favorable textures, and it is not a compulsory choice for our SelfDeblur to adopt another non-blind deconvolution method to generate final deblurring result.

%Given the blurry image $\bf y$, the optimization of Eqn. \eqref{eq:unconstrained neural blind deconvolution} can also be explained as a kind of "zero-shot" self-supervised learning \cite{shocher2018zero} of $\mathcal{G}_k$ and $\mathcal{G}_x$, and our proposed method is dubbed SelfDeblur.

\begin{algorithm}[!t] \small
	\caption{SelfDeblur (Alternating Optimization)}
	\label{algm:self deblur ADM}
	\begin{algorithmic}[1]
		\Require Blurry image $\mathbf{y}$
		\Ensure Blur kernel $\mathbf{k}$ and clean image $\mathbf{x}$
		\State Sample $\mathbf{z}_x$ and $\mathbf{z}_k$ from uniform distribution with seed 0.
		\State $\mathbf{k} = \mathcal{G}_k^0(\mathbf{z}_k)$
		\For{$t = 1 \text{ to } T$}
		
		\State $\mathbf{x} = \mathcal{G}^{t-1}_x(\mathbf{z}_x)$
		\State Compute the gradient w.r.t. $\mathcal{G}_k$
		\State Update $\mathcal{G}_k^t$ using the ADAM algorithm \cite{kingma2014adam}
		\State $\mathbf{k} = \mathcal{G}^{t}_k(\mathbf{z}_k)$
		\State Compute the gradient w.r.t. $\mathcal{G}_x$
		\State Update $\mathcal{G}_x^t$ using the ADAM algorithm \cite{kingma2014adam}
		
		\EndFor
		\State    $\mathbf{x} =  \mathcal{G}^{T}_x(\mathbf{z}_x)$, $\mathbf{k} =  \mathcal{G}^{T}_x(\mathbf{z}_k)$
	\end{algorithmic}
\end{algorithm}

\begin{algorithm}[!ht] \small
	\caption{SelfDeblur (Joint Optimization)}
	\label{algm:self deblur}
	\begin{algorithmic}[1]
		\Require Blurry image $\mathbf{y}$
		\Ensure Blur kernel $\mathbf{k}$ and clean image $\mathbf{x}$
		\State Sample $\mathbf{z}_x$ and $\mathbf{z}_k$ from uniform distribution with seed 0.
		\For{$t = 1 \text{ to } T$}
		
		\State $\mathbf{k} = \mathcal{G}^{t-1}_k(\mathbf{z}_k)$
		\State $\mathbf{x} = \mathcal{G}_x^{t-1}(\mathbf{z}_x)$
		\State Compute the gradients w.r.t. $\mathcal{G}_k$ and $\mathcal{G}_x$
		\State Update $\mathcal{G}_k^t$ and $\mathcal{G}_x^t$ using the ADAM algorithm \cite{kingma2014adam}
		
		\EndFor
		\State   $\mathbf{x} =  \mathcal{G}^{T}_x(\mathbf{z}_x)$, $\mathbf{k} =  \mathcal{G}^{T}_x(\mathbf{z}_k)$
	
	\end{algorithmic}

\end{algorithm}

\vspace{-0.1in}
\section{Experimental Results}
\vspace{-0.1in}
In this section, ablation study is first conducted to analyze the effect of optimization algorithm and network architecture.
Then, our SelfDeblur is evaluated on two benchmark datasets and is compared with the state-of-the-art blind deconvolution methods.
Finally, we report the results of SelfDeblur on several real-world blurry images.

Our SelfDeblur is implemented using Pytorch \cite{paszke2017automatic}. %, and is optimized using ADAM \cite{kingma2014adam} algorithm.
The experiments are conducted on a PC equipped with one NVIDIA Titan V GPU.
Unless specially stated, the experiments follow the same settings, \ie, $T=5,000$, and the noises $\mathbf{z}_x$ and $\mathbf{z}_k$ are sampled from the uniform distribution with fixed random seed 0.
Following~\cite{ulyanov2018deep}, we further perturb $\mathbf{z}_x$ randomly at each iteration.
The initial learning rate is set as $0.01$ and is decayed by multiplying 0.5 when reaching 2,000, 3,000 and 4,000 iterations.
%
%The source code and trained models will be publicly available. 

\begin{figure*}[!htb]
	\vspace{-0.2in}
	\setlength{\abovecaptionskip}{0.cm}
	\setlength{\belowcaptionskip}{-0.cm}
	\centering
	%\fbox{\rule{0pt}{3in} \rule{0.9\linewidth}{0pt}}
	\setlength{\tabcolsep}{1pt}
	\begin{tabular}{ccccccc}
		\!\!\!\!\!
		\includegraphics[width=.36\textwidth]{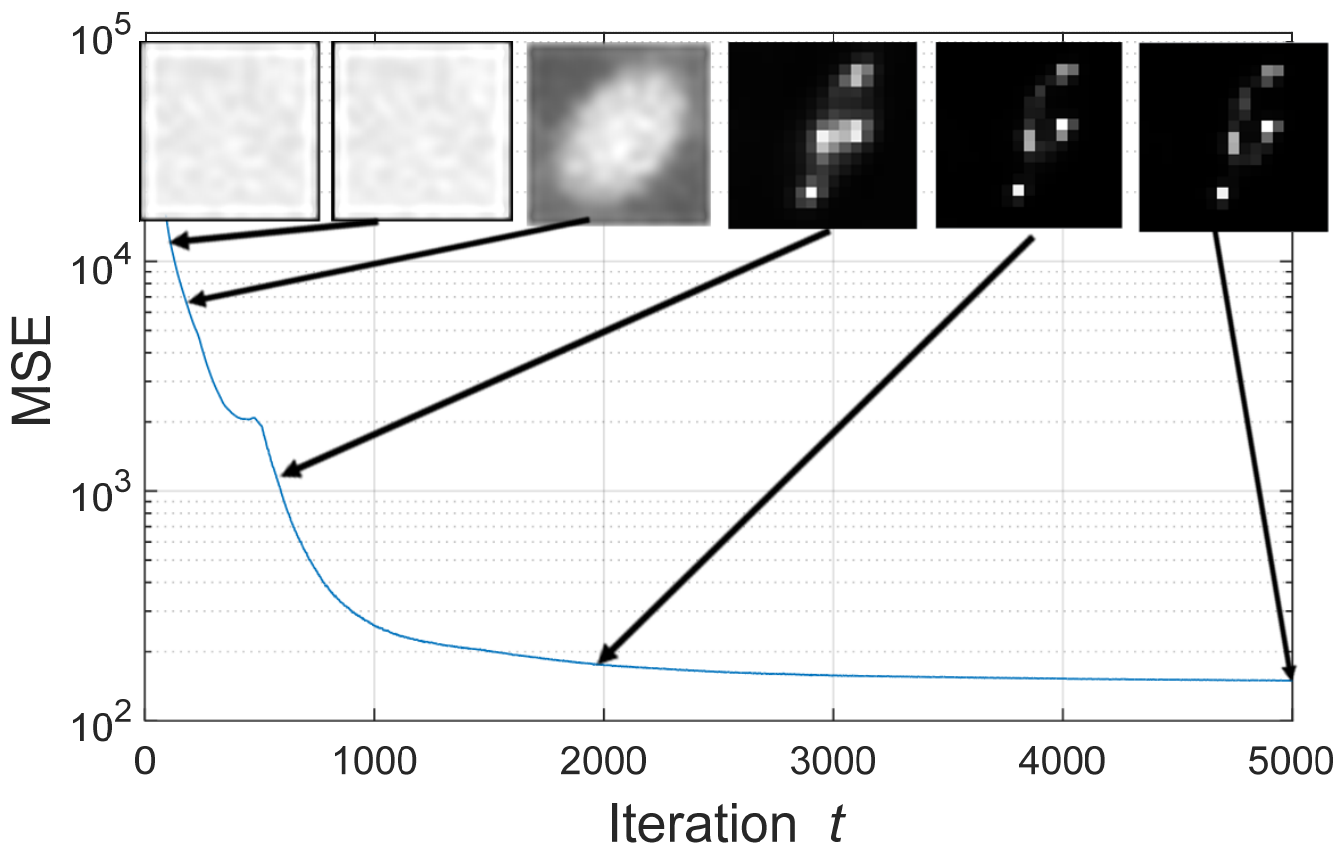}\ \ 
		\includegraphics[width=.36\textwidth]{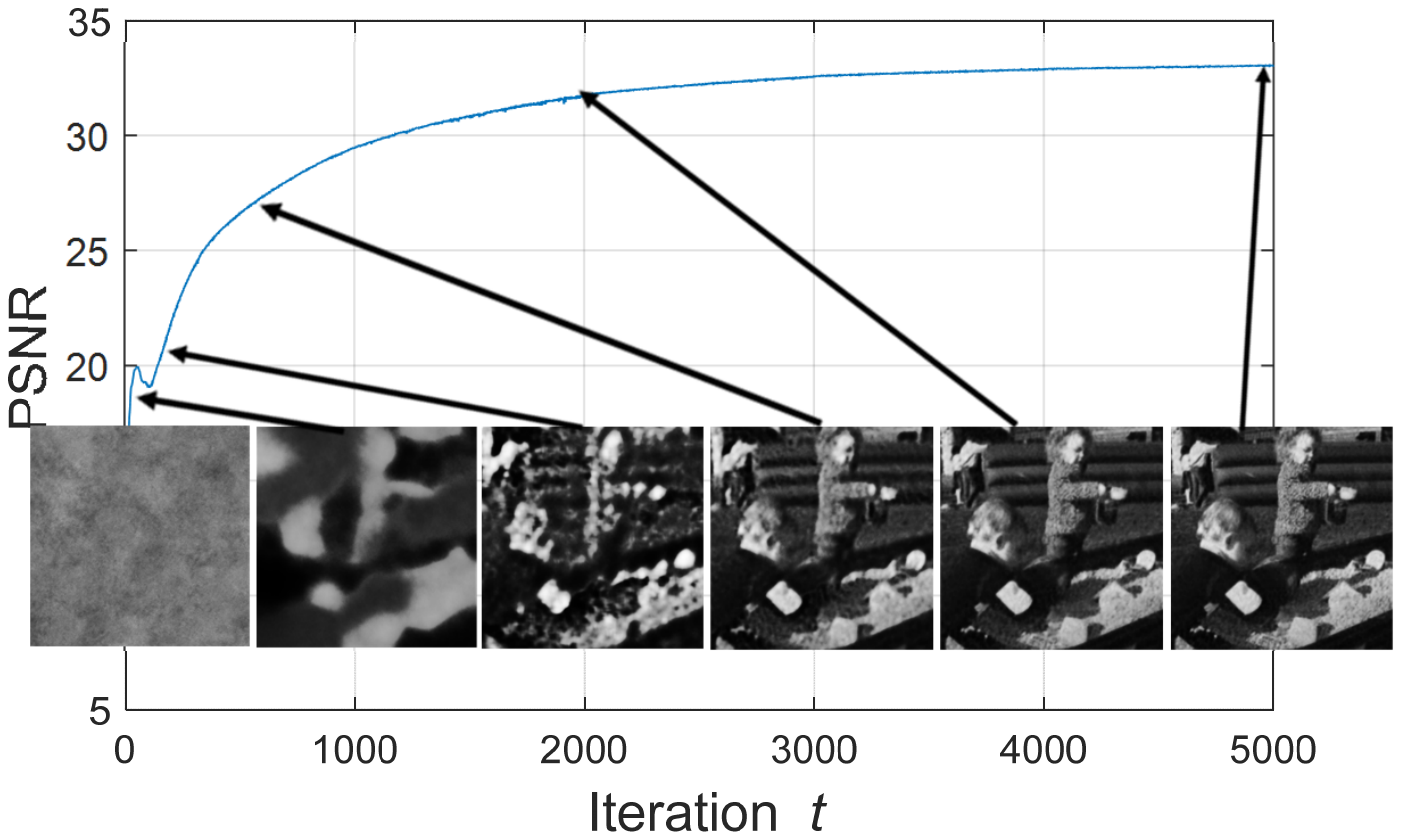}\\
		%		MSE of $\bf k$ & PSNR of $\bf x$  \\
	\end{tabular}
	\caption{Intermediate results of estimated blur kernel and latent clean image at iteration $t=1, 20, 100, 600, 2,000 \text{ and } 5,000$. }
	\label{fig:intermediate}
	\vspace{-0.25in}
\end{figure*}

%\vspace{-0.1in}
\subsection{Ablation Study}
%\vspace{-0.1in}

Ablation study is conducted on the dataset by Levin \etal \cite{levin2009understanding}, which is a popular blind deconvolution benchmark consisting of 4 clean images and 8 blur kernels.
%
%The blurry images in Levin \etal's dataset are with low noise levels, and thus Levin \etal's dataset is an ideal testing bed to conduct ablation studies of SelfDeblur.
%
Using \cite{zoran2009scale}, the average estimated noise level of the blurry images in the dataset is $\sigma \approx 1\times 10^{-5}$.
Thus we simply adopt $\lambda=1\times10^{-6}$ on this dataset.

\vspace{-0.1in}
\subsubsection{Alternating Optimization \emph{vs}. Joint Optimization} \label{sec:experiment joint}
\vspace{-0.1in}

We first evaluate the performance of SelfDeblur using alternating optimization (SelfDeblur-A) and joint optimization (SelfDeblur-J).
%
%On Levin \etal's dataset \cite{levin2009understanding}, SelfDeblur-A and SelfDeblur-J run with the same settings, \ie, $T=5,000$ and $\lambda=10^{-6}$.
%
Table \ref{table:ablation} reports the average PSNR and SSIM values.
In terms of quantitative metrics, SelfDeblur-J significantly outperforms SelfDeblur-A, demonstrating the superiority of joint optimization.
In the supplementary file, we provide several failure cases of SelfDeblur-A, where SelfDeblur-A may converge to delta kernel and worse solution while SelfDeblur-J performs favorably on these cases.
Therefore, joint optimization is adopted as the default SelfDeblur method throughout the following experiments.

\begin{table}[!htbp] \footnotesize
	\vspace{-0.1in}
	\centering
	\setlength{\abovecaptionskip}{0.cm}
	\setlength{\belowcaptionskip}{-0.cm}
	\caption{Average PSNR/SSIM comparison of SelfDeblur-A and SelfDeblur-J on the dataset of Levin \etal \cite{levin2009understanding}.
	} \label{table:ablation}
	\setlength{\tabcolsep}{25pt}
	\begin{tabular}{c|cccc}
		\hline
		SelfDeblur-A& SelfDeblur-J  \\
		\hline
		30.53 / 0.8748 & 33.07 / 0.9313 \\
		\hline
	\end{tabular}
\vspace{-0.2in}
\end{table}

\vspace{-0.1in}
\subsubsection{Network Architecture of $\mathcal{G}_k$}\label{sec:experiment architecture}
\vspace{-0.1in}
In this experiment, we compare the results by considering four kinds of network architectures: (i) SelfDeblur, (ii) Double-DIP~\cite{gandelsman2019double} (asymmetric Autoencoder with skip connections for both $\mathcal{G}_x$ and $\mathcal{G}_k$), (iii) SelfDeblur$_{k-}$ (removing the hidden layer from $\mathcal{G}_k$), and (iv) SelfDeblur$_{k+}$ (adding an extra hidden layer for $\mathcal{G}_k$).
%
%Similar to Double-DIP~\cite{gandelsman2019double}, the deep networks for generating natural images can be directly adopted in SelfDeblur.
%
%We have adopted Autorencoder with skip connections in DIP [39] and progressive generator (PG) in PGGAN \cite{karras2018progressive} to serve as both $\mathcal{G}_x$ and $\mathcal{G}_k$ (namely SKIP$^2$ and PG$^2$).
%
From Table~\ref{table:selfdeblur variants} and Fig.~\ref{fig:variant results}, SelfDeblur significantly outperforms Double-DIP in estimating blur kernel and latent image.
The result indicates that the DIP network is limited to capture the prior of blur kernel, and the simple FCN can be a good choice of $\mathcal{G}_k$.
%
%Even PG is suggested to exploit multi-scale information, PG$^2$ tends to converge to delta kernel solution.
%
%SKIP is effective in image generation, but usually generates coarse blur kernels.
%
%Considering the simplicity of $\bf k$, the simple 1D FCN is a good choice for $\mathcal{G}_k$.
%
We further compare SelfDeblur with SelfDeblur$_{k-}$ and SelfDeblur$_{k+}$.
One can see that the FCN without hidden layer (\ie, SelfDeblur$_{k-}$) also succeeds in estimating blur kernel and clean image (see Fig.~\ref{fig:variant results}), but performs much inferior to SelfDeblur.
Moreover, the three-layer FCN (\ie, SelfDeblur$_{k+}$) is superior to SelfDeblur$_{k-}$, but is inferior to SelfDeblur. 
To sum up, SelfDeblur is a good choice for modeling blur kernel prior.

\begin{table}[!htb]\footnotesize
	\vspace{-0.1in}
	\setlength{\abovecaptionskip}{0.cm}
	\setlength{\belowcaptionskip}{-0.cm}
	\caption{Quantitavie comparison of SelfDeblur variants with different network structures of $\mathcal{G}_k$.}
	\setlength{\tabcolsep}{5pt}
	\centering
	
	\begin{tabular}{c|cccccccc}
		\hline
		
		\hline
		& SelfDeblur  & SelfDeblur$_{k-}$ & SelfDeblur$_{k+}$ & Double-DIP   \\
		\hline
		PSNR    &33.07   &  28.37 &30.92  & 21.51          \\
		SSIM    &0.9313 & 0.8396 &0.8889  & 0.5256       \\
		\hline
		
		\hline
	\end{tabular}
	\label{table:selfdeblur variants}
	\vspace{-0.2in}
\end{table}

\begin{figure}[!htb]\footnotesize
	\setlength{\abovecaptionskip}{0.cm}
	\setlength{\belowcaptionskip}{-0.cm}
	\centering
	%\fbox{\rule{0pt}{3in} \rule{0.9\linewidth}{0pt}}
	\setlength{\tabcolsep}{1pt}
	\begin{tabular}{ccccccc}
		
		\includegraphics[width=.1\textwidth]{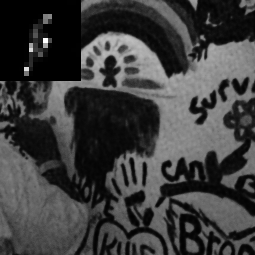} &
		\includegraphics[width=.1\textwidth]{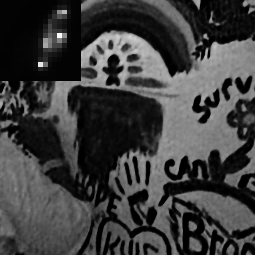} &
		\includegraphics[width=.1\textwidth]{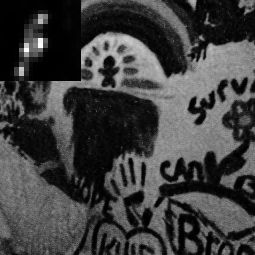}&
		\includegraphics[width=.1\textwidth]{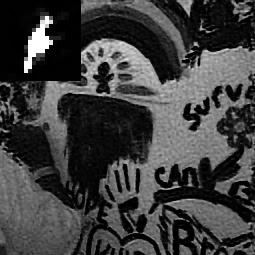} \\
		SelfDeblur & SelfDeblur$_{k-}$ & SelfDeblur$_{k+}$ & Double-DIP  \\
	\end{tabular}
	\caption{Visual comparison of SelfDeblur variants with different network structures of $\mathcal{G}_k$. }
	\label{fig:variant results}
	\vspace{-0.2in}
\end{figure}

\vspace{-0.1in}
\subsubsection{Visualization of Intermediate Results}
\vspace{-0.1in}
Using an image from the dataset of Levin \etal~\cite{levin2009understanding}, Fig.~\ref{fig:intermediate} shows the intermediate results of estimated blur kernel and clean image at iteration $t$ $=$ $1, 20, 100, 600, 2,000 \text{ and } 5,000$, along with the MSE curve for $\bf k$ and the PSNR curve for $\bf x$.
%
%We visulize the intermediate results of SelfDeblur on an example from Levin \etal's dataset \cite{levin2009understanding}.
%To show the power of the deep priors, TV regularizaion is removed on this case, \ie, $\lambda=0$ in Eqn. \eqref{eq:selfdeblur function}.
%
%Fig.~\ref{fig:intermediate} shows intermediate blur kernels and clean images at iteration $t$ $=$ $1, 20, 100, 600, 2,000 \text{ and } 5,000$, along with MSE curve for $\bf k$ and PSNR curve for $\bf x$.
%
When iteration $t=20$, the intermediate result of $\bf x$ mainly contains the salient image structures, which is consistent with the observation that salient edges is crucial for initial blur kernel estimation in traditional methods.
%
%and then contributes to coarse estimation of $\bf k$, which is consistent with the crucial edge prediction in traditional blind deconvolution methods.
%
%In traditional methods, time-varying parameters are tuned to gradually identify and preserve salient edges, and texrue details in latter iterations would still destroy the convergence of blur kernel estimation.
%
Along with the increase of iterations, $\mathcal{G}_x$ and $\mathcal{G}_k$ begin to generate finer details in $\bf x$ and $\bf k$.
Unlike traditional methods, SelfDeblur is effective in simultaneously estimating blur kernel and recovering latent clean image when iteration $t \geq 20$, making the non-blind deconvolution not a compulsory choice for SelfDeblur.

%It is interesting to see that the deep priors in SelfDeblur are sufficient to simultaneously estimate blur kernel and recover latent clean image with fine details.

\begin{figure*}[!htb]\footnotesize
	\centering
	\setlength{\tabcolsep}{0pt}
	\setlength{\abovecaptionskip}{0.cm}
	\setlength{\belowcaptionskip}{-0.cm}
	\begin{tabular}{cclcclcclcclcclccl}
		
		\multicolumn{3}{c}{\includegraphics[width=.21\textwidth]{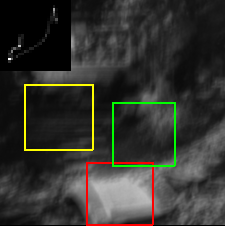}}\ &
		\multicolumn{3}{c}{\includegraphics[width=.21\textwidth]{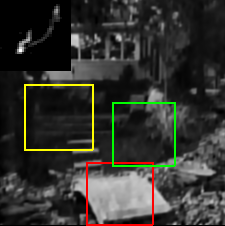}}\ &
		\multicolumn{3}{c}{\includegraphics[width=.21\textwidth]{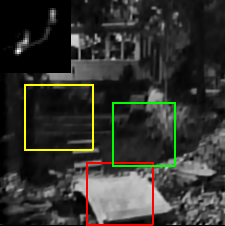}}\ &
		\multicolumn{3}{c}{\includegraphics[width=.21\textwidth]{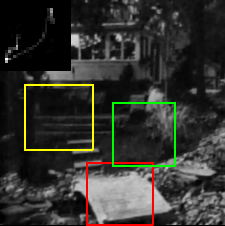}}\vspace{-2pt}\\
		\includegraphics[width=.07\textwidth]{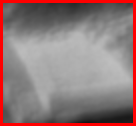} &
		\includegraphics[width=.07\textwidth]{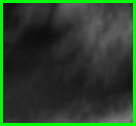} &
		\includegraphics[width=.07\textwidth]{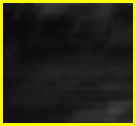}\ &
		\includegraphics[width=.07\textwidth]{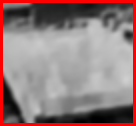} &
		\includegraphics[width=.07\textwidth]{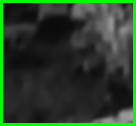} &
		\includegraphics[width=.07\textwidth]{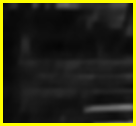}\ &
		\includegraphics[width=.07\textwidth]{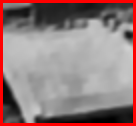} &
		\includegraphics[width=.07\textwidth]{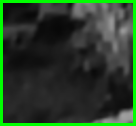} &
		\includegraphics[width=.07\textwidth]{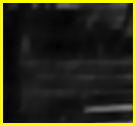}\ &
		\includegraphics[width=.07\textwidth]{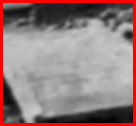} &
		\includegraphics[width=.07\textwidth]{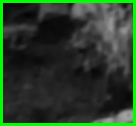} &
		\includegraphics[width=.07\textwidth]{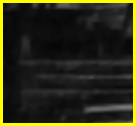}\\
		\multicolumn{3}{c}{Blurry image} &
		\multicolumn{3}{c}{Zuo \etal$^\Delta$\cite{zuo2016learning}} &
		\multicolumn{3}{c}{Xu\&Jia$^\Delta$\cite{xu2010two}} &
		\multicolumn{3}{c}{SelfDeblur$^\Delta$}\\
		\multicolumn{3}{c}{\includegraphics[width=.21\textwidth]{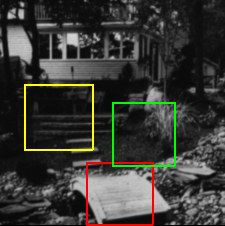}}\ &
		\multicolumn{3}{c}{\includegraphics[width=.21\textwidth]{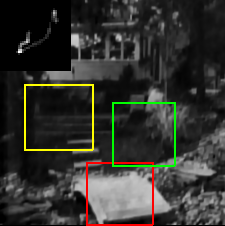}}\ &
		\multicolumn{3}{c}{\includegraphics[width=.21\textwidth]{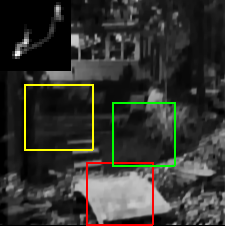}}\ &
		\multicolumn{3}{c}{\includegraphics[width=.21\textwidth]{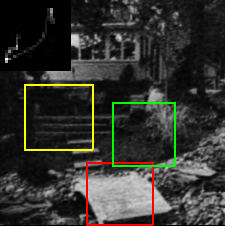}}\vspace{-2pt}\\
		\includegraphics[width=.07\textwidth]{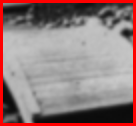} &
		\includegraphics[width=.07\textwidth]{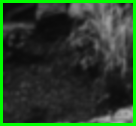} &
		\includegraphics[width=.07\textwidth]{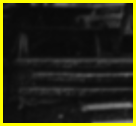}\ &
		\includegraphics[width=.07\textwidth]{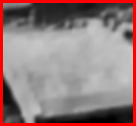} &
		\includegraphics[width=.07\textwidth]{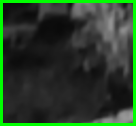} &
		\includegraphics[width=.07\textwidth]{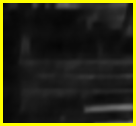}\ &
		\includegraphics[width=.07\textwidth]{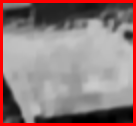} &
		\includegraphics[width=.07\textwidth]{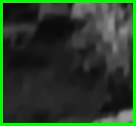} &
		\includegraphics[width=.07\textwidth]{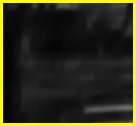}\ &
		\includegraphics[width=.07\textwidth]{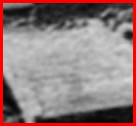} &
		\includegraphics[width=.07\textwidth]{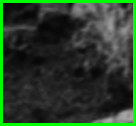} &
		\includegraphics[width=.07\textwidth]{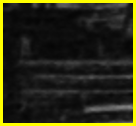}\\
		\multicolumn{3}{c}{Ground-truth} &
		\multicolumn{3}{c}{Sun \etal$^\Delta$\cite{sun2013edge}} &
		\multicolumn{3}{c}{Pan-DCP$^\Delta$\cite{pan2018deblurring}} &
		\multicolumn{3}{c}{SelfDeblur}\\
		
	\end{tabular}
	\caption{Visual comparison on the dataset of Levin \etal \cite{levin2009understanding}.
	}
	\label{fig:levin results}
	\vspace{-0.1in}
\end{figure*}

%\vspace{-0.05in}
\subsection{Comparison with State-of-the-arts}
\vspace{-0.05in}

\subsubsection{Results on dataset of Levin \etal \cite{levin2009understanding}}
\vspace{-0.1in}
\begin{table}[htbp] \footnotesize
	\vspace{-0.1in}
	\setlength{\abovecaptionskip}{0.cm}
	\setlength{\belowcaptionskip}{-0.cm}
	\caption{Average PSNR, SSIM, Error Ratio and running time (sec.) comparison on the dataset of Levin \etal \cite{levin2009understanding}.
		$^\Delta$ indicates the method generates final deblurring results using the non-blind deconvolution method from \cite{levin2011efficient}.
		The running time only includes blur kernel estimation.} \label{table:levin}
	\centering
	\setlength{\tabcolsep}{7pt}
	\begin{tabular}{c|ccccccc}
		\hline
		& PSNR & SSIM & Error Ratio  & Time\\
		\hline
		Known \textbf{k}$^\Delta$	& 34.53 &	0.9492 &	1.0000 & --- \\
		Krishnan \etal$^\Delta$ \cite{krishnan2011blind} & 29.88 &	0.8666 &	2.4523 & 8.9400 \\
		Cho\&Lee$^\Delta$ \cite{cho2009fast}&30.57&	0.8966&	1.7113 & 1.3951\\
		Levin \etal$^\Delta$ \cite{levin2011efficient}&30.80&	0.9092&	1.7724 & 78.263\\
		Xu\&Jia$^\Delta$ \cite{xu2013unnatural}&31.67&	0.9163&	1.4898 & 1.1840\\
		Sun \etal$^\Delta$ \cite{sun2013edge}&32.99&	0.9330&	1.2847 & 191.03\\
		Zuo \etal$^\Delta$ \cite{zuo2016learning}&	32.66&	0.9332&	1.2500 & 10.998\\
		Pan-DCP$^\Delta$ \cite{pan2018deblurring} & 32.69 &0.9284 &1.2555 &295.23 \\
		SRN\cite{tao2018scale} & 23.43 &0.7117 &6.0864 & N/A\\
		\hline
		SelfDeblur$^\Delta$& 33.32 &0.9438 &1.2509 & --- \\
		SelfDeblur &33.07 & 0.9313 & 1.1968 & 224.01\\
		\hline
	\end{tabular}
	\vspace{-0.1in}
\end{table}
\begin{table*}[!htb]\footnotesize
	\setlength{\abovecaptionskip}{0.cm}
	\setlength{\belowcaptionskip}{-0.cm}
	\caption{Average PSNR/SSIM comparison on the dataset of Lai \etal \cite{lai2016comparative}, which has 5 categories.
		The methods marked with $^\Delta$ adopt \cite{whyte2014deblurring} and \cite{krishnan2009fast} as non-blind deconvolution after blur kernel estimation in \emph{Saturated} and the other categories, respectively.
	}
	\centering
	\setlength{\tabcolsep}{1pt}
	\begin{tabular}{c|ccccccc|ccc}
		\hline
		
		\hline
		Images &  Cho\&Lee$^\Delta$\cite{cho2009fast} & Xu\&Jia$^\Delta$\cite{xu2010two} &  Xu \etal$^\Delta$\cite{xu2013unnatural}     & Michaeli \etal$^\Delta$\cite{michaeli2014blind} & Perroe \etal$^\Delta$ \cite{perrone2014total} & Pan-L0$^\Delta$\cite{pan2017l_0} & Pan-DCP$^\Delta$\cite{pan2018deblurring} & SelfDeblur$^\Delta$ & SelfDeblur\\
		\hline
		\emph{Manmade}   &16.35/0.3890    & 19.23/0.6540    &17.99/0.5986    & 17.43/0.4189    &17.41/0.5507 &16.92/0.5316&18.59/0.5942&20.08/0.7338 &20.35/0.7543    \\
		\emph{Natural}   &20.14/0.5198    &23.03/0.7542     &21.58/0.6788    &20.70/0.5116     &21.04/0.6764 &20.92/0.6622&22.60/0.6984&22.50/0.7183&22.05/0.7092  \\
		\emph{People} & 19.90/0.5560   & 25.32/0.8517    & 24.40/0.8133   &  23.35/0.6999   &22.77/0.7347 &23.36/0.7822&24.03/0.7719&27.41/0.8784&25.94/0.8834\\
		\emph{Saturated} &14.05/0.4927    & 14.79/0.5632    & 14.53/0.5383   & 14.14/0.4914    &14.24/0.5107 &14.62/0.5451&16.52/0.6322&16.58/0.6165&16.35/0.6364\\
		\emph{Text} & 14.87/0.4429   &18.56/0.7171     & 17.64/0.6677   & 16.23/0.4686    &16.94/0.5927 &16.87/0.6030&17.42/0.6193&19.06/0.7126&20.16/0.7785\\
		\hline
		\textbf{Avg}. &17.06/0.4801 &20.18/0.7080 &19.23/0.6593 & 18.37/0.5181&18.48/0.6130 &18.54/0.6248 & 19.89/0.6656& 21.13/0.7319 &  20.97/0.7524\\
		\hline
		
		\hline
	\end{tabular}
	\label{table:cvpr16 dataset}
	\vspace{-0.2in}
\end{table*}
Using the dataset of Levin \etal \cite{levin2009understanding}, we compare our SelfDeblur  with several state-of-the-art blind deconvolution methods, including Krishnan \etal \cite{krishnan2011blind}, Levin \etal \cite{levin2009understanding}, Cho\&Lee \cite{cho2009fast}, Xu\&Jia \cite{xu2010two}, Sun \etal \cite{sun2013edge}, Zuo \etal \cite{zuo2016learning} and Pan-DCP \cite{pan2018deblurring}.
Besides, SelfDeblur is compared with one state-of-the-art deep motion deblurring method SRN \cite{tao2018scale}, which is re-trained on 1,600 blurry images \cite{ren2019simultaneous} synthesized using eight blur kernels in the dataset of Levin \etal.  
For SelfDeblur, $\lambda = 1\times 10^{-6}$ is set for all the blurry images.
Following \cite{sun2013edge,zuo2016learning}, we adopt the non-blind deconvolution method in \cite{levin2011efficient} to generate final deblurring results.
PSNR, SSIM \cite{wang2004image} and Error Ratio \cite{levin2011efficient} are used as quantitative metrics.
And we also report the running time of blur kernel estimation for each competing method.
Our SelfDeblur and SRN are ran on an NVIDIA Titan V GPU, while the other methods are ran on a PC with 3.30GHz Intel(R) Xeon(R) CPU.
%

%, which is defined as
%\begin{equation}\label{eq:error ratio}
%\text{Error Ratio} = \frac{\|\mathbf{x}-\mathbf{x}^{gt}\|}{\|\bar{\mathbf{x}}-\mathbf{x}^{gt}\|}	
%\end{equation}
%where $\mathbf{x}$ is the deblurring result by blind deconvolution method, $\bar{\mathbf{x}}$ is the deblurring result given ground-truth blur kernel, and $\mathbf{x}^{gt}$ is the ground-truth clean image.

\begin{figure*}[!htb]\footnotesize
	\centering
	\setlength{\tabcolsep}{0pt}
	\setlength{\abovecaptionskip}{0.cm}
	\setlength{\belowcaptionskip}{-0.cm}
	\begin{tabular}{cclcclcclcclcclccl}
		\multicolumn{3}{c}{\includegraphics[width=.21\textwidth]{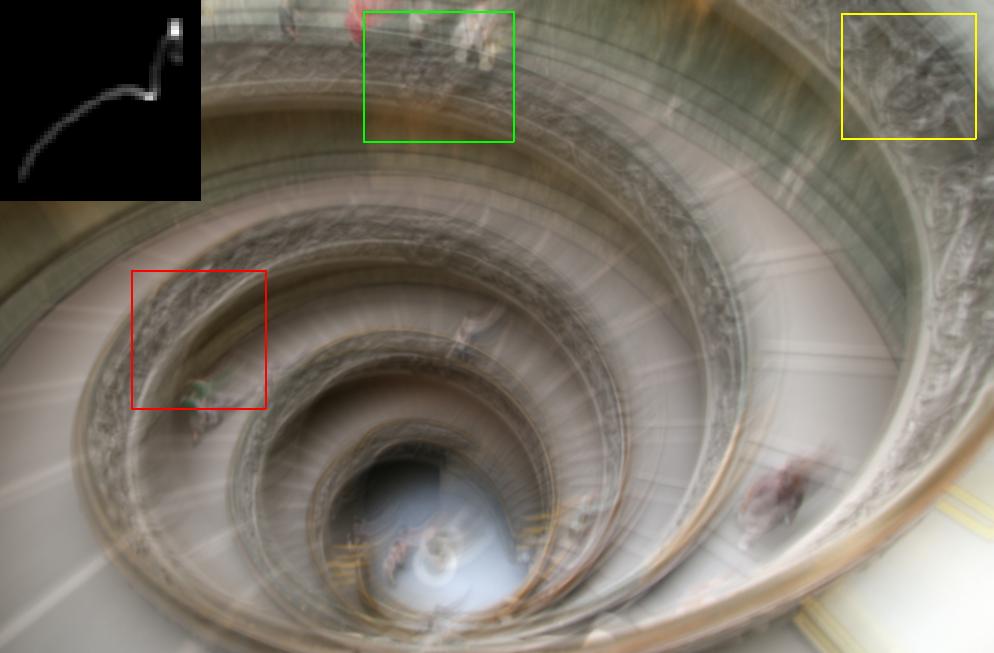}}\ &
		\multicolumn{3}{c}{\includegraphics[width=.21\textwidth]{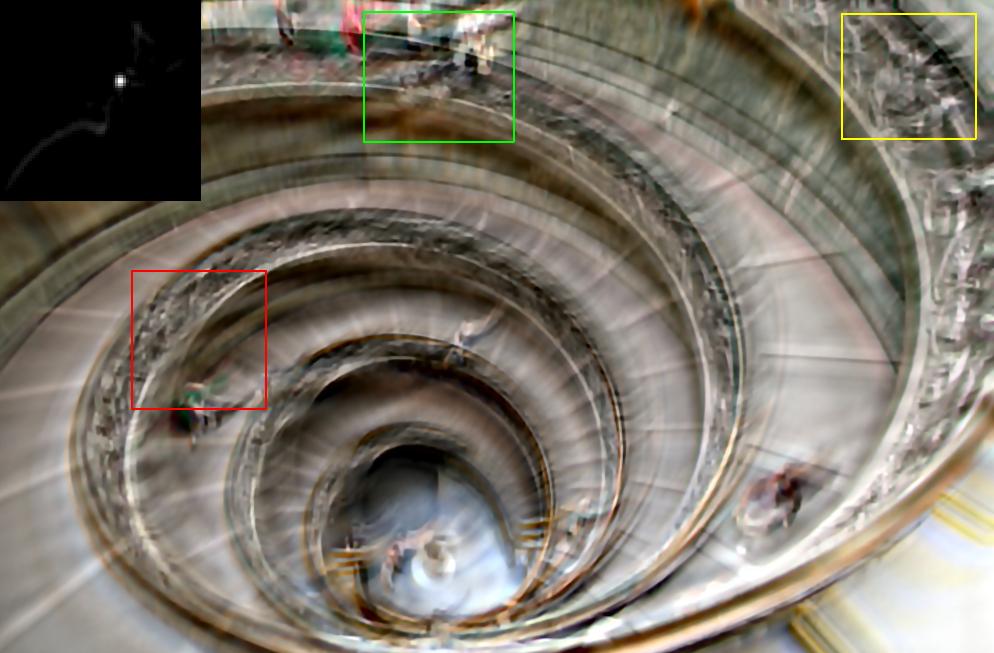}}\ &
		\multicolumn{3}{c}{\includegraphics[width=.21\textwidth]{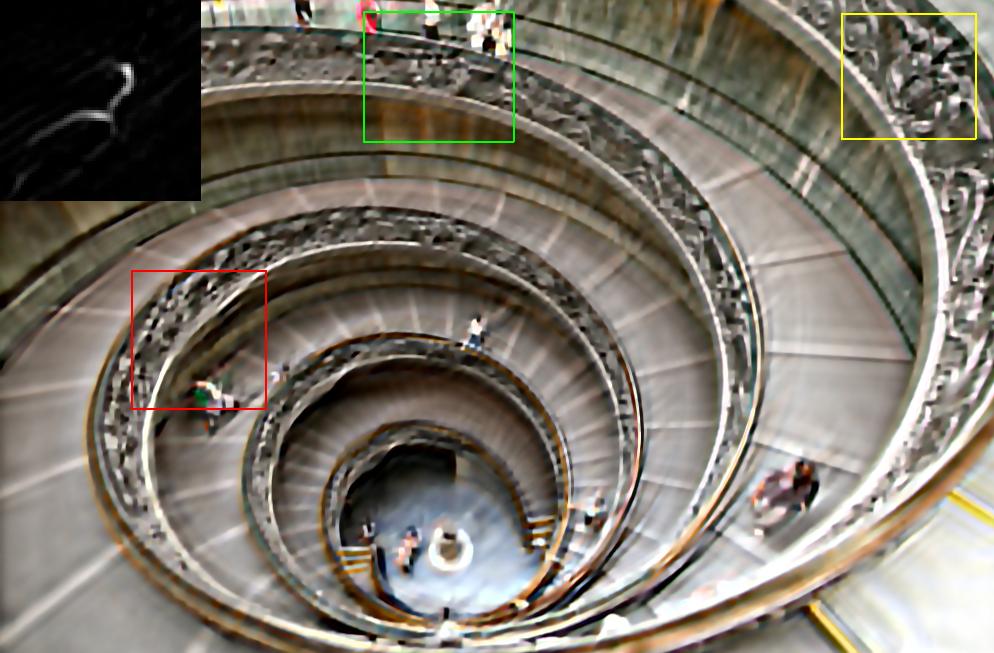}}\ &
		\multicolumn{3}{c}{\includegraphics[width=.21\textwidth]{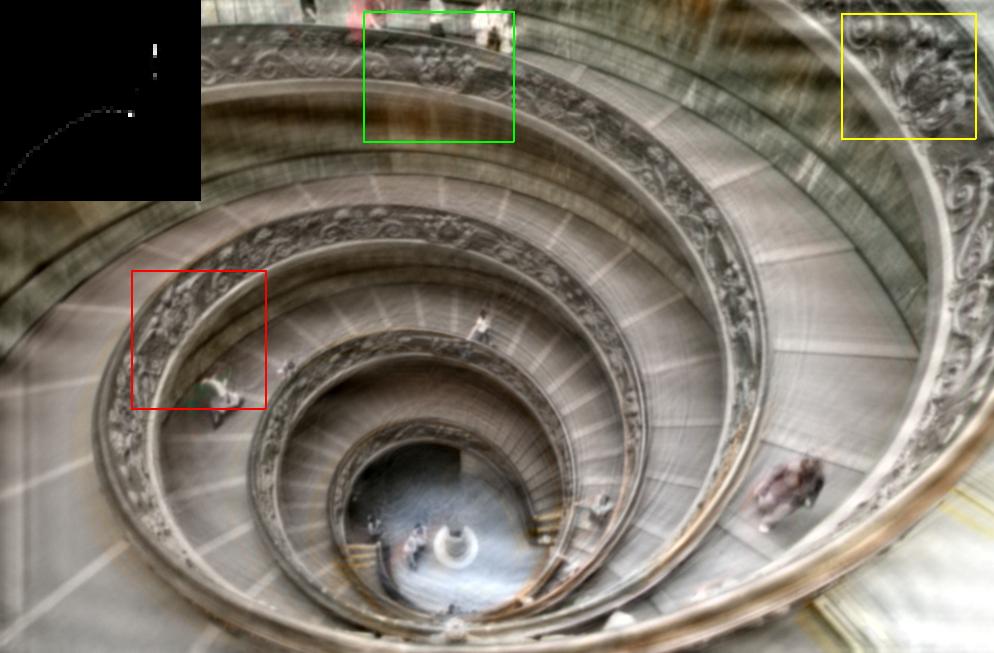}}\vspace{-2pt}\\
		\includegraphics[width=.07\textwidth]{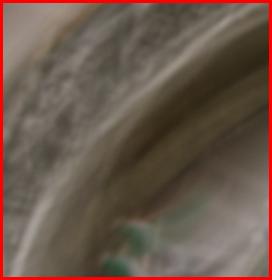} &
		\includegraphics[width=.07\textwidth]{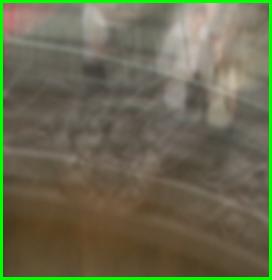} &
		\includegraphics[width=.07\textwidth]{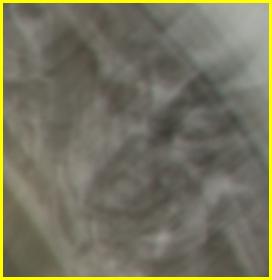}\ &
		\includegraphics[width=.07\textwidth]{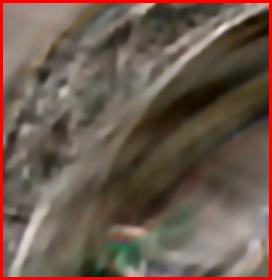} &
		\includegraphics[width=.07\textwidth]{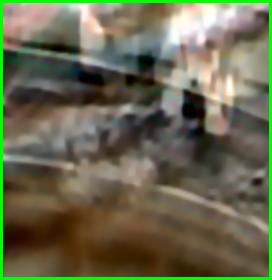} &
		\includegraphics[width=.07\textwidth]{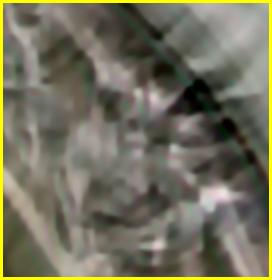}\ &
		\includegraphics[width=.07\textwidth]{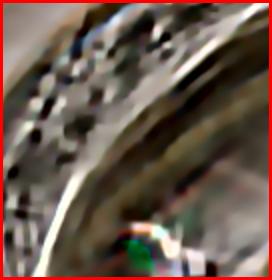} &
		\includegraphics[width=.07\textwidth]{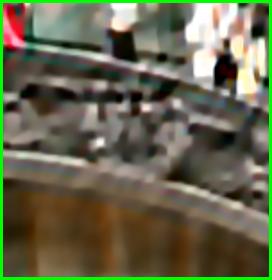} &
		\includegraphics[width=.07\textwidth]{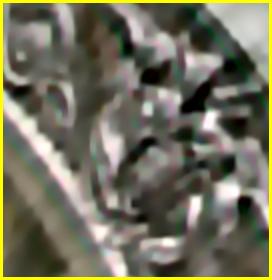}\ &
		\includegraphics[width=.07\textwidth]{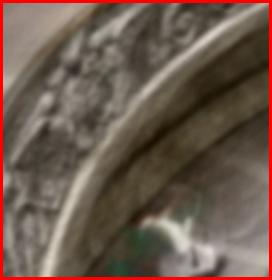} &
		\includegraphics[width=.07\textwidth]{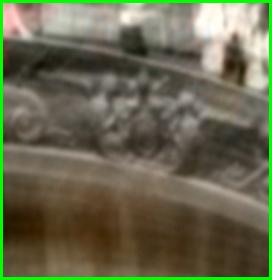} &
		\includegraphics[width=.07\textwidth]{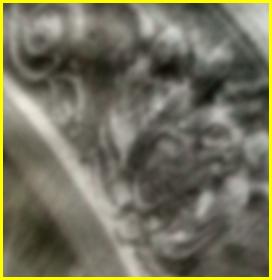}\\
		\multicolumn{3}{c}{Blurry image} &
		\multicolumn{3}{c}{Xu\&Jia$^\Delta$\cite{xu2010two}} &
		\multicolumn{3}{c}{Perrone \etal$^\Delta$\cite{perrone2014total}} &
		\multicolumn{3}{c}{SelfDeblur$^\Delta$}\\
		\multicolumn{3}{c}{\includegraphics[width=.21\textwidth]{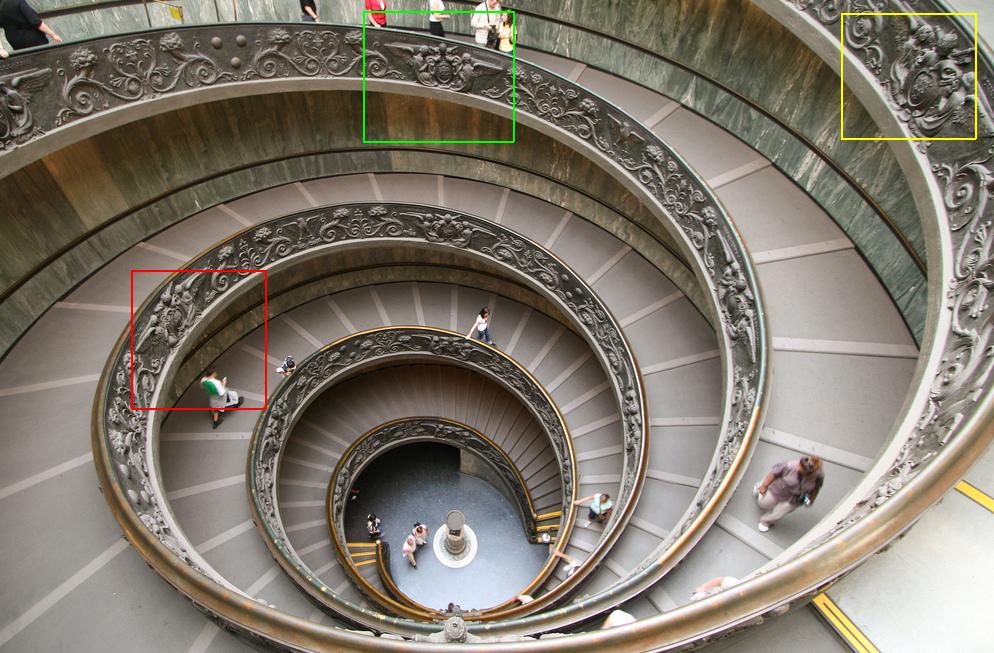}}\ &
		\multicolumn{3}{c}{\includegraphics[width=.21\textwidth]{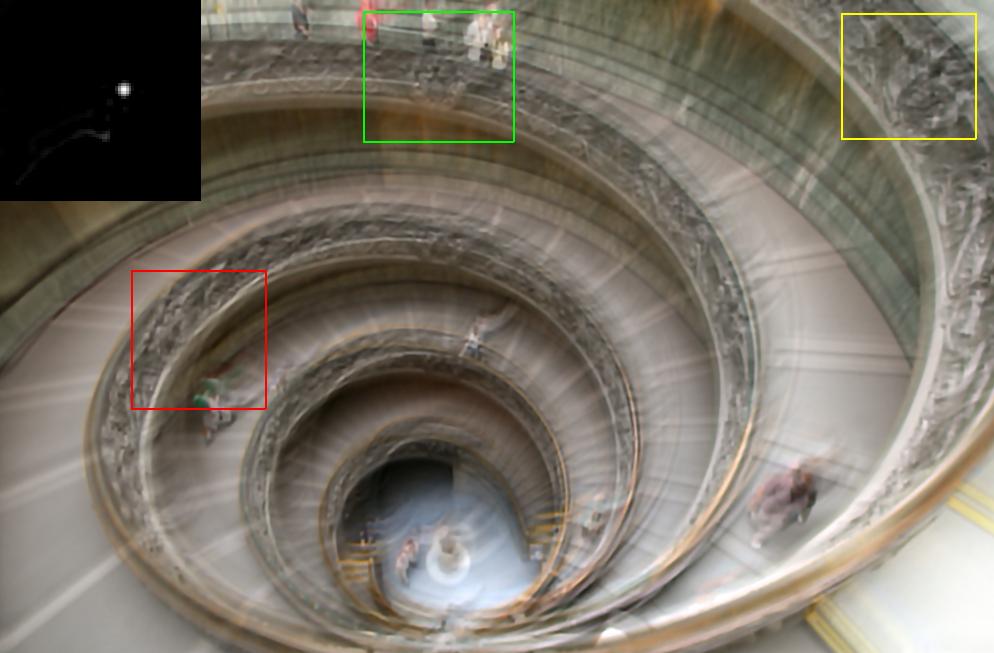}}\ &
		\multicolumn{3}{c}{\includegraphics[width=.21\textwidth]{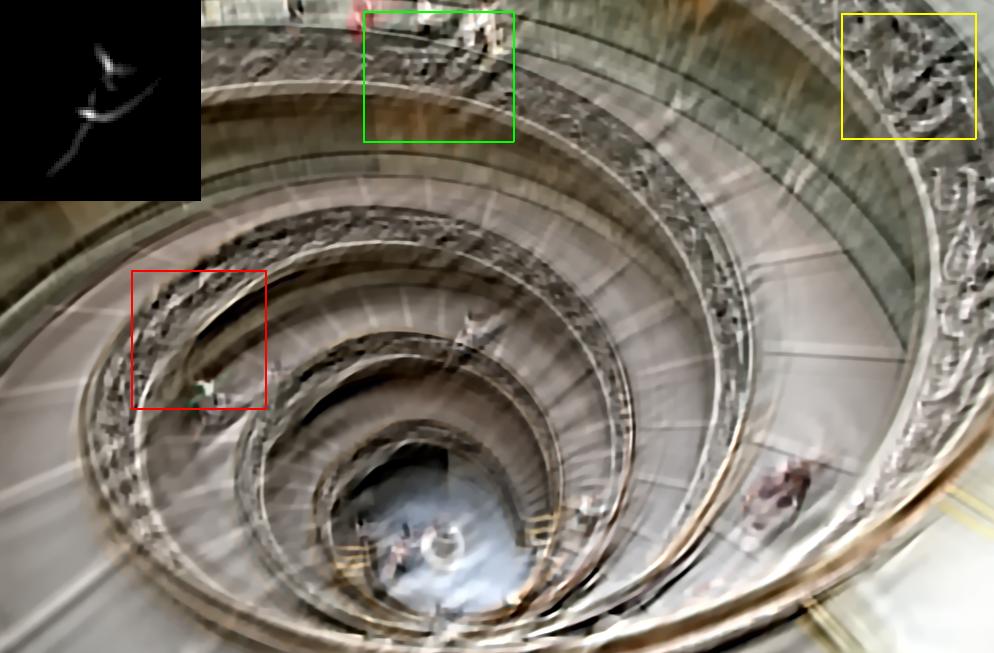}}\ &
		\multicolumn{3}{c}{\includegraphics[width=.21\textwidth]{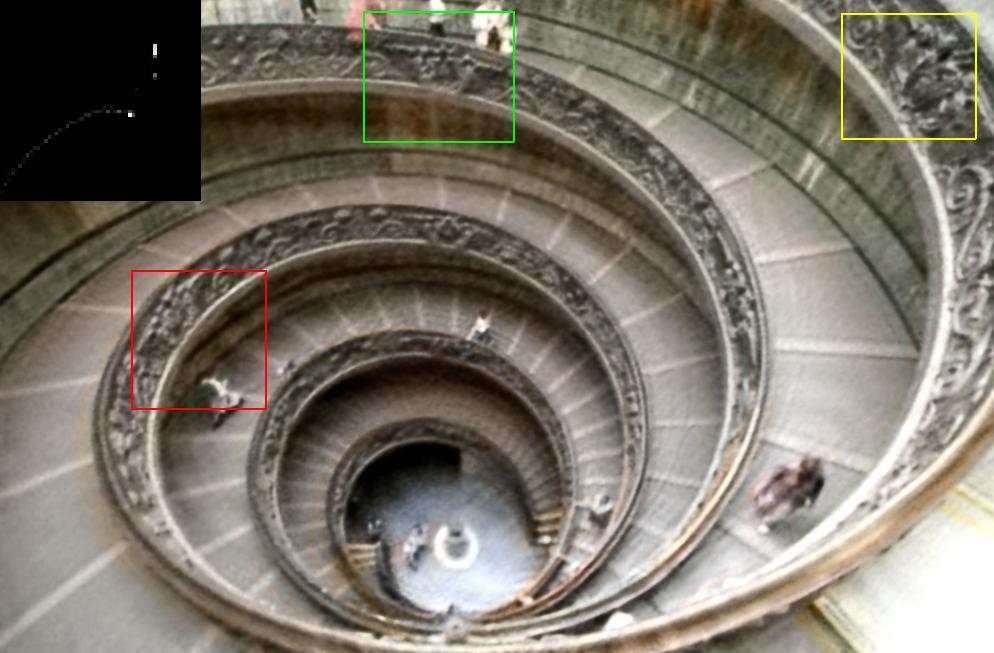}}\vspace{-2pt}\\
		\includegraphics[width=.07\textwidth]{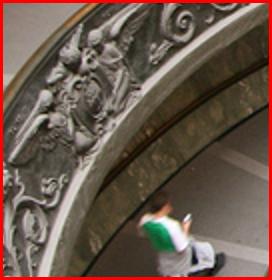} &
		\includegraphics[width=.07\textwidth]{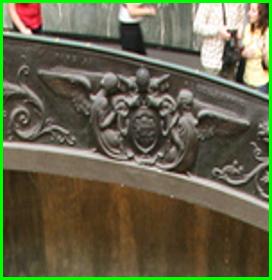} &
		\includegraphics[width=.07\textwidth]{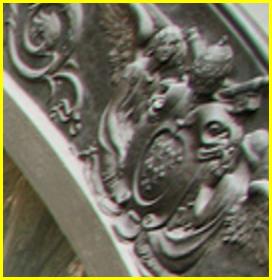}\ &
		\includegraphics[width=.07\textwidth]{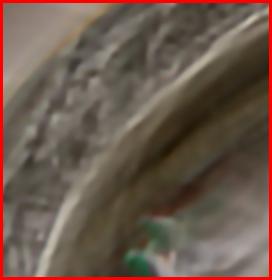} &
		\includegraphics[width=.07\textwidth]{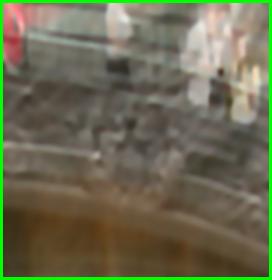} &
		\includegraphics[width=.07\textwidth]{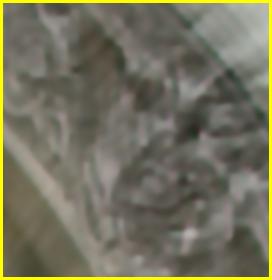}\ &
		\includegraphics[width=.07\textwidth]{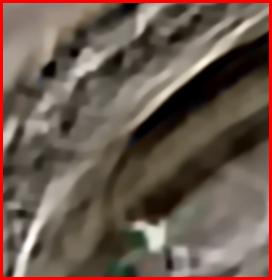} &
		\includegraphics[width=.07\textwidth]{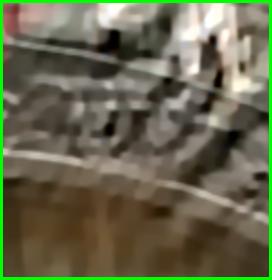} &
		\includegraphics[width=.07\textwidth]{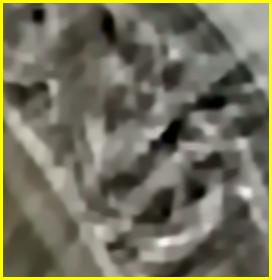}\ &
		\includegraphics[width=.07\textwidth]{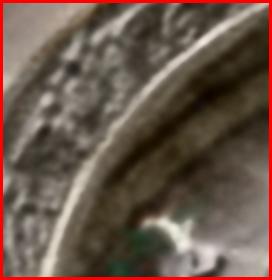} &
		\includegraphics[width=.07\textwidth]{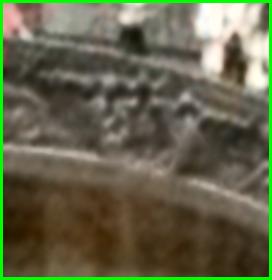} &
		\includegraphics[width=.07\textwidth]{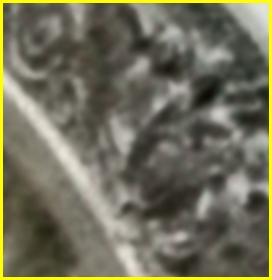}\\
		\multicolumn{3}{c}{Ground-truth} &
		\multicolumn{3}{c}{Michaeli \etal$^\Delta$\cite{xu2010two}} &
		\multicolumn{3}{c}{Pan-DCP$^\Delta$\cite{perrone2014total}} &
		\multicolumn{3}{c}{SelfDeblur}\\
		
	\end{tabular}
	\caption{Visual comparison on the dataset of Lai \etal \cite{lai2016comparative}.}
	\label{fig:cvpr16 results}
	\vspace{-0.1in}
\end{figure*}

\begin{figure*}[!htb]\footnotesize
	\centering
	\setlength{\tabcolsep}{0pt}
	\setlength{\abovecaptionskip}{0.cm}
	\setlength{\belowcaptionskip}{-0.cm}
	\begin{tabular}{cclcclcclcclcclccl}
		
		\multicolumn{3}{c}{\includegraphics[width=.21\textwidth]{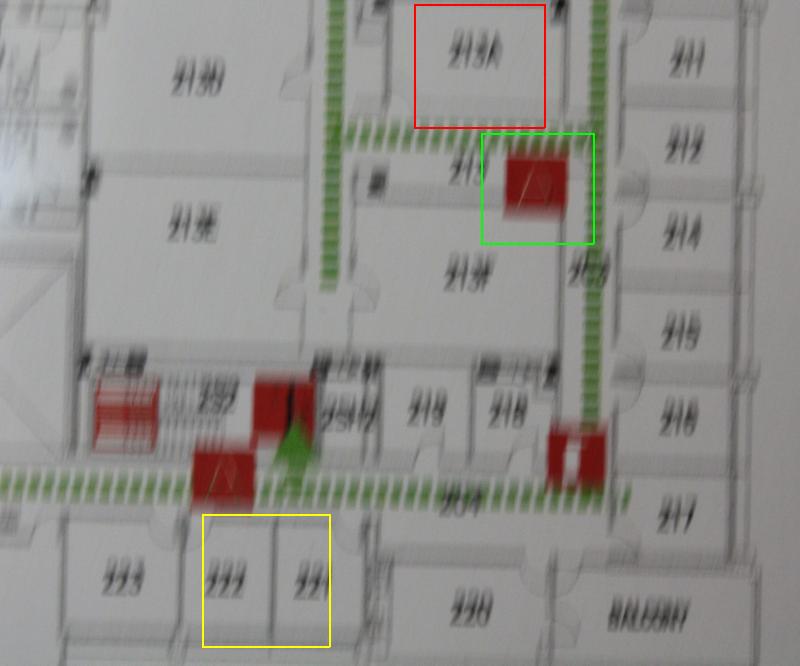}}\ &
		\multicolumn{3}{c}{\includegraphics[width=.21\textwidth]{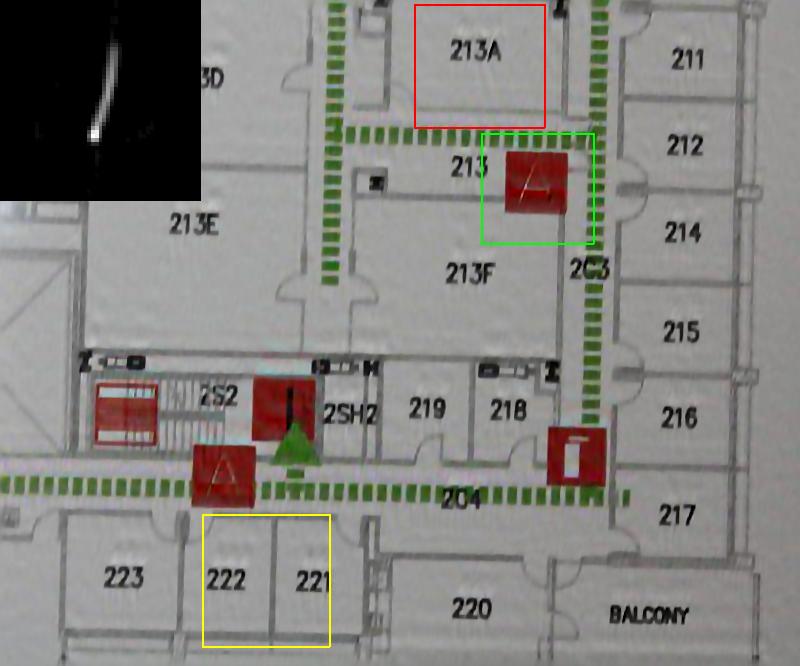}}\ &
		\multicolumn{3}{c}{\includegraphics[width=.21\textwidth]{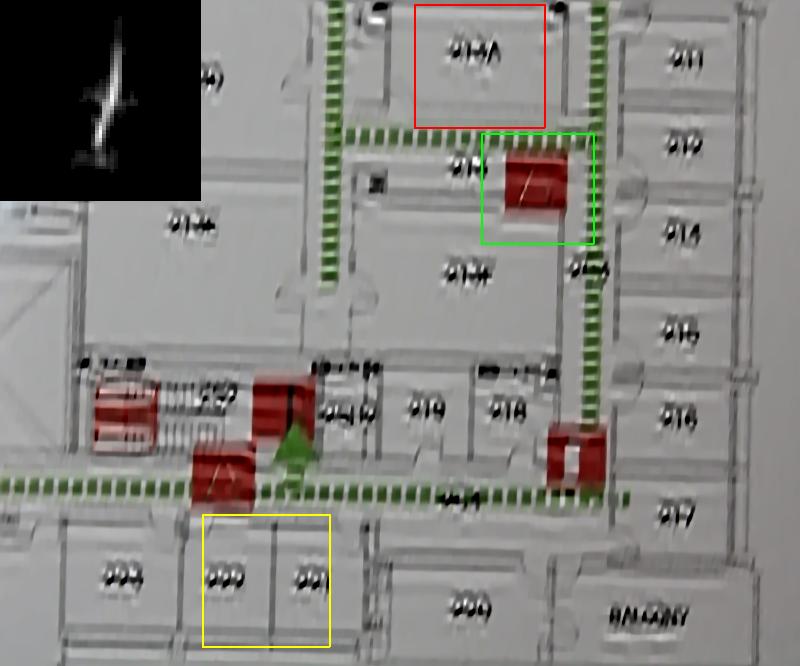}}\ &
		\multicolumn{3}{c}{\includegraphics[width=.21\textwidth]{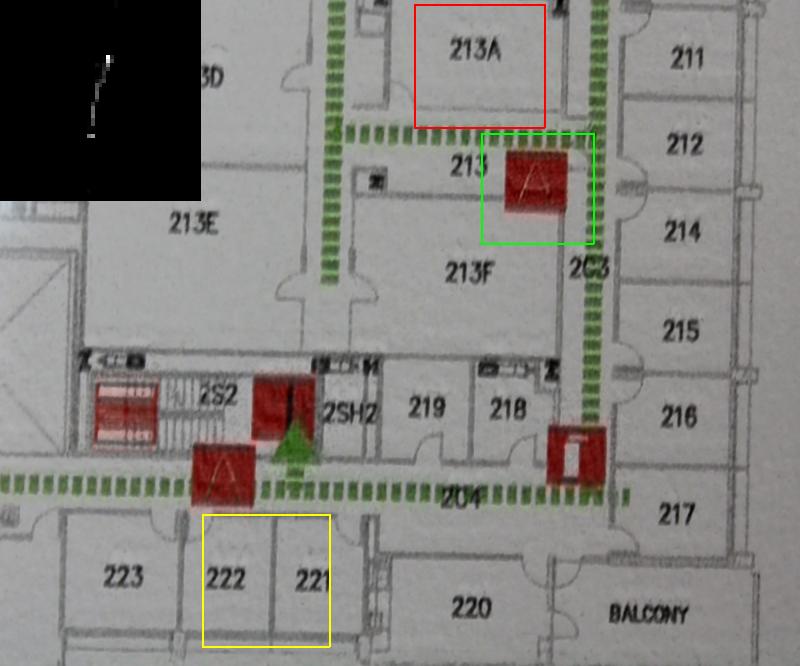}}\vspace{-2pt}\\
		\includegraphics[width=.07\textwidth]{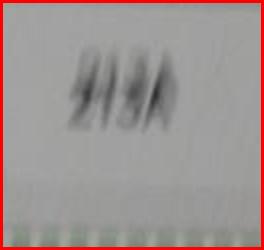} &
		\includegraphics[width=.07\textwidth]{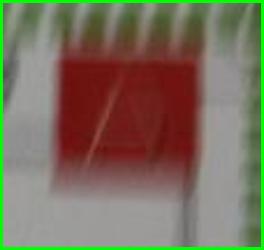} &
		\includegraphics[width=.07\textwidth]{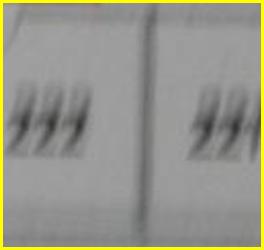}\ &
		\includegraphics[width=.07\textwidth]{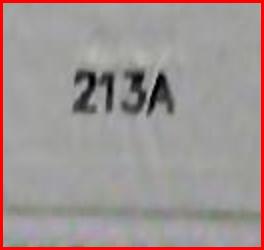} &
		\includegraphics[width=.07\textwidth]{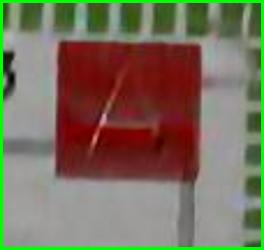} &
		\includegraphics[width=.07\textwidth]{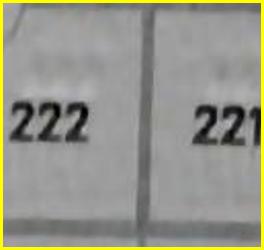}\ &
		\includegraphics[width=.07\textwidth]{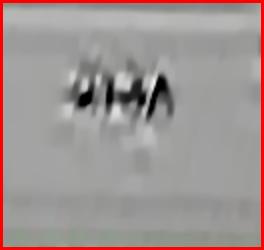} &
		\includegraphics[width=.07\textwidth]{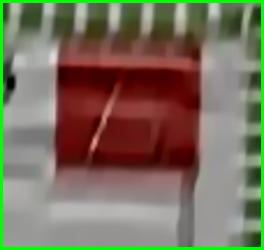} &
		\includegraphics[width=.07\textwidth]{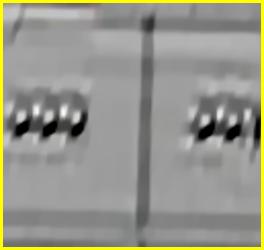}\ &
		\includegraphics[width=.07\textwidth]{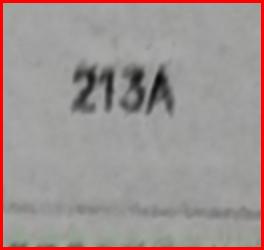} &
		\includegraphics[width=.07\textwidth]{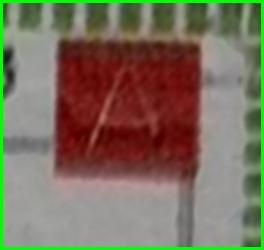} &
		\includegraphics[width=.07\textwidth]{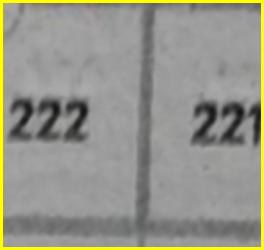}\\
%		\multicolumn{3}{c}{\includegraphics[width=.21\textwidth]{figs/real/ian1/blurry_im}}\ &
%		\multicolumn{3}{c}{\includegraphics[width=.21\textwidth]{figs/real/ian1/xujia_im}}\ &
%		\multicolumn{3}{c}{\includegraphics[width=.21\textwidth]{figs/real/ian1/dcp_im}}\ &
%		\multicolumn{3}{c}{\includegraphics[width=.21\textwidth]{figs/real/ian1/selfdeblur_im}}\vspace{-2pt}\\
%		\includegraphics[width=.07\textwidth]{figs/real/ian1/blurry1} &
%		\includegraphics[width=.07\textwidth]{figs/real/ian1/blurry2} &
%		\includegraphics[width=.07\textwidth]{figs/real/ian1/blurry3}\ &
%		\includegraphics[width=.07\textwidth]{figs/real/ian1/xujia1} &
%		\includegraphics[width=.07\textwidth]{figs/real/ian1/xujia2} &
%		\includegraphics[width=.07\textwidth]{figs/real/ian1/xujia3}\ &
%		\includegraphics[width=.07\textwidth]{figs/real/ian1/dcp1} &
%		\includegraphics[width=.07\textwidth]{figs/real/ian1/dcp2} &
%		\includegraphics[width=.07\textwidth]{figs/real/ian1/dcp3}\ &
%		\includegraphics[width=.07\textwidth]{figs/real/ian1/selfdeblur1} &
%		\includegraphics[width=.07\textwidth]{figs/real/ian1/selfdeblur2} &
%		\includegraphics[width=.07\textwidth]{figs/real/ian1/selfdeblur3}\\
		\multicolumn{3}{c}{Blurry image} &
		\multicolumn{3}{c}{Xu\&Jia\cite{xu2010two}} &
		\multicolumn{3}{c}{Pan-DCP\cite{pan2018deblurring}} &
		\multicolumn{3}{c}{SelfDeblur}\\
		
	\end{tabular}
	\caption{Visual comparison on two real-world blurry images. }
	\label{fig:real results}
	\vspace{-0.2in}
\end{figure*}

Table \ref{table:levin} lists the average metrics of the competing methods.
We report the results of SelfDeblur with two settings, \ie, the deblurring results purely by SelfDeblur and those using the non-blind deconvolution from \cite{levin2011efficient}, denoted as SelfDeblur$^\Delta$.
In terms of PSNR and Error Ratio, SelfDeblur significantly outperforms the competing methods.
As for average SSIM, SelfDeblur performs slightly inferior to Sun \etal and Zuo \etal.
%
%This is because the deblurring results by our SelfDeblur are with finer texture details than the other methods, while the SSIM metric usually favors smooth images.
%
By incorporating with non-blind deconvolution from \cite{levin2011efficient}, SelfDeblur$^\Delta$ can further boost quantitative performance and outperforms all the other methods.
In terms of running time, SelfDeblur is time-consuming due to the optimization of two generative networks, but is comparable with Sun \etal \cite{sun2013edge} and Pan-DCP \cite{pan2018deblurring}.
From the visual comparison in Fig.~\ref{fig:levin results}, the \#4 blur kernel estimated by SelfDeblur is much closer to the ground-truth.
As shown in the close-ups, SelfDeblur and SelfDeblur$^\Delta$ can recover more visually favorable textures.
We also note that both the performance gap and visual quality between SelfDeblur and SelfDeblur$^\Delta$ are not significant, and thus non-blind deconvolution is not a compulsory choice for our SelfDeblur.

%Especially, the results by SelfDeblur has more detailed textures due to the deep priors.
%
%Thus in practice, our SelfDeblur is not compulsory to perform non-blind deconvolution.

\vspace{-0.1in}
\subsubsection{Results on dataset of Lai \etal \cite{lai2016comparative}}
%\vspace{-0.1in}
%
We further evaluate SelfDeblur on the dataset of Lai \etal \cite{lai2016comparative} consisting of 25 clean images and 4 large size blur kernels.
The blurry images are divided into five categories, \ie, \emph{Manmade}, \emph{Natural}, \emph{People}, \emph{Saturated} and \emph{Text}, where each category contains 20 blurry images.
For each blurry image, the parameter $\lambda$ is set according to the noise level estimated using \cite{zoran2009scale}.
We compare our SelfDeblur with Cho\&Lee \cite{cho2009fast}, Xu\&Jia\cite{xu2010two}, Xu \etal \cite{xu2013unnatural}, Machaeli \etal \cite{michaeli2014blind}, Perroe \etal \cite{perrone2014total}, Pan-L$0$ \cite{pan2017l_0} and Pan-DCP \cite{pan2018deblurring}.
The results of competing methods except Pan-DCP \cite{pan2018deblurring} and ours are duplicated from \cite{lai2016comparative}.
The results of Pan-DCP \cite{pan2018deblurring} are generated using their default settings.
Once the blur kernel is estimated, non-blind deconvolution \cite{krishnan2009fast} is applied to the images of \emph{Manmade}, \emph{Natural}, \emph{People} and \emph{Text}, while \cite{whyte2014deblurring} is used to handle \emph{Saturated} images.
From Table \ref{table:cvpr16 dataset}, both SelfDeblur and SelfDeblur$^\Delta$ can achieve better quantitative metrics than the competing methods.
In terms of image contents, our SelfDeblur outperforms the other methods on any of the five categories.
%
%The size of blur kernels from \#1 to \#4 increases.
%
From the results in Fig. \ref{fig:cvpr16 results}, the blur kernel estimated by our SelfDeblur is more accurate than those by the competing methods, and the deconvolution result is with more visually plausible textures.

\vspace{-0.06in}
\subsection{Evaluation on Real-world Blurry Images}
\vspace{-0.08in}
Our SelfDeblur is further compared with Xu\&Jia \cite{xu2010two} and Pan-DCP \cite{pan2018deblurring} on real-world blurry images.
From Fig.~\ref{fig:real results}, one can see that the blur kernel estimated by our SelfDeblur contains less noises, and the estimated clean image is with more visually plausible structures and textures.
The kernel estimation errors by Xu\&Jia and Pan-DCP are obvious, thereby yielding ringing artifacts in the estimated clean images.
More results can be found in Suppl.

\vspace{-0.1in}
\section{Conclusion}
\vspace{-0.1in}
In this paper, we proposed a neural blind deconvolution method, \ie, SelfDeblur.
It adopts an asymmetric Autoencoder and a FCN to respectively capture the deep priors of latent clean image and blur kernel.
And the \emph{SoftMax} nonlinearity is applied to the output of FCN to meet the non-negative and equality constraints of blur kernel.
A joint optimization algorithm is suggested to solve the unconstrained neural blind deconvolution model.
%
%Additionally, the deep priors are effective in generating latent clean image with visually favorable textures.
%
Experiments show that our SelfDeblur achieves notable performance gains over the state-of-the-art methods, and is effective in estimating blur kernel and generating clean image with visually favorable textures.
%
%The future works can be carried out to address several issues of SelfDeblur \eg, computational efficiency, the extension of deep priors to general cases, \etc.

\vspace{-0.03in}
\section*{Acknowledgements}
\vspace{-0.03in}
This work was supported by the National Natural Science Foundation of China under Grants (Nos. 61801326, 61671182, 61732011 and 61925602), the SenseTime Research Fund for Young Scholars, and the Innovation Foundation of Tianjin University.
%We also gratefully acknowledge
%the support of NVIDIA Corporation with the donation
%of GPU.

\clearpage

{\small
	\bibliographystyle{ieee}
	\bibliography{reference}
}

\end{document}